\documentclass[10pt]{article}

\usepackage[margin=1in]{geometry}
\usepackage{lmodern}
\usepackage[T1]{fontenc}
\usepackage{amsmath}
\usepackage{amssymb}
\usepackage{graphicx}
\usepackage{booktabs}
\usepackage{multirow}
\usepackage{threeparttable}
\usepackage{float}
\usepackage{caption}
\usepackage{enumitem}
\usepackage{xcolor}
\usepackage{microtype}
\usepackage[numbers,sort&compress]{natbib}
\usepackage[colorlinks=true,linkcolor=blue!50!black,citecolor=blue!50!black,urlcolor=blue!50!black,
  pdftitle={Token Reduction Is Not Cost Reduction: An Empirical Study of End-to-End Efficiency in API-Based Coding Agents},
  pdfauthor={Sarel Weinberger, Amir Hozez},
  pdfsubject={Empirical study of token and cost efficiency in API-based coding agents},
  pdfkeywords={coding agents, prompt caching, context compression, retrieval, cost efficiency, benchmarks}]{hyperref}
\usepackage{cleveref}

\setlist{itemsep=1pt,topsep=3pt,parsep=0pt}

\newcommand{\ninegate}{RTK-ML}
\newcommand{\Ninegate}{RTK-ML}
\newcommand{\ci}[2]{[#1,\,#2]}

\title{\textbf{Token Reduction Is Not Cost Reduction:\\
An Empirical Study of End-to-End Efficiency\\ in API-Based Coding Agents}}

\author{%
  Sarel~Weinberger\\
  \small PointFive\\
  \small \texttt{sarel.weinberger@pointfive.co}%
\and
  Amir~Hozez\\
  \small PointFive\\
  \small \texttt{amir@pointfive.co}%
}
\date{July 2026}

\begin{document}
\maketitle

\begin{abstract}
Token-reduction tools for coding agents are often evaluated by the number of
tokens they remove, but token count alone does not determine end-to-end
inference cost. We evaluate context- and tool-output reduction in Claude Code
using paired, provider-billed coding tasks, measuring task success, cache
traffic, agent behavior, and cost. Across the experiments, tool-output
reduction was not a reliable predictor of lower billed cost. The most
aggressive compression setup reduced delivered tool-output tokens by 38.4\%
but increased billed cost by 6.8\%. Cost decomposition shows that prompt-cache
traffic accounts for most measured input-side cost, limiting the fraction of
spend directly addressable by tool-output compression. We also find that
compression can alter the agent trajectory by inducing additional retrieval,
diagnosis, testing, or turns, offsetting local token savings. In a
SWE-bench-derived Go subset, aggressive compression also reduced task
completion. These results show that token reduction alone is an inadequate
proxy for cost reduction in tool-heavy coding agents. Efficiency should
therefore be evaluated using success-adjusted end-to-end cost, including cache
behavior, trajectory changes, and correctness.
\end{abstract}

\section{Introduction}\label{sec:intro}

API-based coding agents such as Claude Code~\citep{anthropic2026claudecode},
SWE-agent~\citep{yang2024sweagent}, and OpenHands~\citep{wang2024openhands}
operate by iteratively invoking tools---shell commands, file reads, searches,
test runs---and feeding the resulting text back into a large language model.
Each turn re-transmits an ever-growing context: repository excerpts, prior
tool outputs, test logs, diffs, and conversation history. This context growth has motivated several optimization approaches:
deterministic command-output compressors, query-aware retrieval rankers,
semantic and structural code search, and API-boundary proxies that rewrite
payloads in flight.

Nearly all of these layers are evaluated by some form of \emph{tokens
removed}: compression ratios on captured command outputs, recall-at-budget on
retrieval corpora, or preserved-marker counts on synthetic fixtures. This study tests whether tokens removed predicts the provider-billed cost of
completing a task successfully in the evaluated workloads. The paired-campaign
results show that token reduction alone is an incomplete predictor of that
outcome.

Three mechanisms invalidate this simple equivalence. First, modern serving stacks
bill cached context at discounted rates~\citep{anthropic2026caching,
gim2024promptcache,kwon2023vllm}; in the workloads we measure, the large
majority of the reconstructed spend is prompt-cache traffic whose price per
token is
10--125\% of the nominal input price, so removing text from it saves
far less than a token count suggests. Second, an agent is a closed loop:
if compression removes context the model later needs, the model can search
again, re-read files, or take extra diagnostic turns, and those added turns
re-transmit the \emph{entire} context prefix. Third, compression can make transformed outputs incompatible with downstream consumers other than the model---in one failure mode we
analyze, ranked search summaries were piped into shell counting pipelines
that expected raw \texttt{grep} lines, producing incorrect answers without an explicit error.

We therefore separate six measurement layers that are frequently conflated:
(1) component-level compression ratios; (2) retrieval-quality metrics;
(3) model-visible context reduction; (4) billed API-token reduction;
(5) end-to-end task success; and (6) cost per successful execution. Using a
completed evidence audit over all retained campaign artifacts (transcripts,
ledgers, manifests, gate traces, and analysis code), we place every system
for which artifacts exist---baseline Claude Code, RTK, the
RTK-ML architecture, Headroom, a vendor CLI compressor, the
\texttt{ml\_lexical} engine, Caveman, and lexical,
embedding-based, structural, and union retrieval---at the highest evidence
layer its artifacts support, and we report end-to-end results only where
paired, actually-billed campaigns exist.

\paragraph{Scale of the broader program.} The results reported here emerged
from a substantially larger research and development effort: across the
broader program, exploratory, superseded, tuning, failure-analysis, and
infrastructure experiments consumed API resources well beyond the campaigns
analyzed here. This paper does not treat that broader spend as a
statistical sample. Instead, it reports in detail the controlled campaigns
for which tasks, configurations, provider-returned usage, success
judgments, transcripts, and analysis artifacts were retained and could be
audited; those auditable campaigns comprise the roughly 5{,}500 billed
executions reconciled in the campaign inventory
(\cref{tab:campaigns}, Appendix~\ref{app:reconciliation}). We disclose the
broader experimental investment to convey the scale of the search process,
while restricting quantitative claims to the smaller evidence set that
satisfies the study's reproducibility and statistical
requirements.\footnote{The broader development expenditure is contextual
and is excluded from all sample sizes, confidence intervals, cost
decompositions, and system comparisons in this paper. No repository ledger
aggregates it, and this paper reports no absolute spend figures.}

\paragraph{Research questions.}
\begin{itemize}
\item \textbf{RQ1.} Which components dominate the billed cost of API-based
  coding agents in realistic paired workloads?
\item \textbf{RQ2.} How much of that cost is addressable by tool-output
  compression and retrieval-context reduction?
\item \textbf{RQ3.} Do component-level compression and retrieval gains
  predict end-to-end task efficiency?
\item \textbf{RQ4.} How do model tier, effort level, task family, and
  trajectory structure mediate system effects?
\item \textbf{RQ5.} What benchmark design is required for trustworthy
  token-efficiency claims?
\end{itemize}

\paragraph{Contributions.}
\begin{enumerate}
\item A calibrated cost decomposition for API-based coding agents: a
  four-component decomposition that reproduces most individual bills with a
  median per-run residual of $\approx$1\%, showing that cache creation and
  reads accounted for $\approx$87\% of the reconstructed cost
  ($\approx$80\% of the actual bill), alongside an explicitly quantified
  8.7\% dollar-weighted residual that the retained telemetry cannot
  attribute (\cref{sec:composition}). A per-component attribution of billed cost
  (\cref{tab:componentshares}) shows that the surfaces a user-side layer
  can modify account for only $\approx$6\% of cost, placing the estimated upper bound for visible-token compression at $\approx$5\% under the measured cost decomposition.
\item A large paired, provider-billed comparison of context-reduction
  layers for coding agents---the primary campaign alone spans
  2,908 runs, 103 tasks, 7 repositories, 3 models, up to 5 effort levels, and
  4 arms from identical fresh working copies with block-randomized order,
  within a measurement program of 5{,}493 billed executions (5{,}123 of
  them in append-only cost ledgers), plus 8{,}263 zero-inference-cost
  component evaluations
  (\cref{sec:method}).
\item Direct evidence that token reduction and cost reduction decouple: a
  38.4\% reduction in estimated raw tool-output tokens coexisting with a $+6.8\%$ paired cost
  increase, and a near-zero per-task correlation ($r=0.15$) between
  tool-output reduction and billed-cost change (\cref{sec:results}).
\item A trajectory- and phase-level account of initial cost reductions and downstream cost increases, built from 13,620 deterministically classified
  assistant turns (\cref{sec:phases}).
\item Two documented compression-safety failure modes---shell-pipeline incompatibility with ranked search output, and alteration of the verbatim edit
  anchors that patch application depends on---together with the mitigations
  that fixed them (\cref{sec:failures,sec:grounded}).
\item A single-shot grounded-completion study on SWE-bench-derived Go
  tasks in which byte-exact context grounding was associated with the
  largest observed solve-rate improvement (1/29 to 5/29), while three tasks
  solved by the grounded raw arm were not solved by the grounded compressed
  arm and no task showed the reverse pattern (exact paired $p=0.25$;
  descriptively higher cost per solve under compression;
  \cref{sec:grounded}).
\item A layered evidence taxonomy and benchmark-design recommendations for
  future token-efficiency claims (\cref{sec:componentgap,sec:recommendations}).
\end{enumerate}

\section{Cost Composition of API-Based Coding Agents}\label{sec:costcomposition}

\subsection{Terminology}\label{sec:terms}

We use the following terms throughout; they are deliberately not
interchangeable.

\begin{description}[leftmargin=1.4em]
\item[Uncached input tokens] Prompt tokens processed at the full input price
  (provider usage field \texttt{input\_tokens}).
\item[Cache-creation tokens] Prompt tokens written into the provider's prompt
  cache (usage field \path{cache_creation_input_tokens}), billed at a
  multiplier of
  the input price.
\item[Cache-read tokens] Prompt tokens served from the prompt cache
  (\texttt{cache\_read\_input\_tokens}), billed at a discount.
\item[Generated output tokens] Tokens produced by the model
  (\texttt{output\_tokens}), including visible text and tool-call arguments.
\item[Tool-use block] A structured tool invocation emitted by the model
  within an assistant turn.
\item[Raw tool output] The bytes a tool actually produced before any
  compression layer (observable in these campaigns only for the RTK arms,
  via the hook-side ledger).
\item[Delivered (model-visible) tool output] The text that reached the model
  after any compression or ranking layer. Raw and delivered tool-output
  token counts are estimates from RTK's embedded local BPE
  tokenizer (\texttt{tiktoken} \texttt{o200k\_base} tables), not the
  provider tokenizer.
\item[Turn] One assistant message; a run's \emph{trajectory length} is its
  number of turns.
\item[Cost per run / per successful execution] Billed dollars per run; and
  \emph{cost per successful execution}: total billed cost across all
  executions in an arm divided by the number of successful executions;
  repeated executions of the same underlying task remain separate attempts.
  We use \emph{success-adjusted cost} for the broader concept, and we do
  not report a per-unique-task estimator.
\end{description}

Raw shell output, delivered tool output, and generated output tokens are
three different quantities; conflating any two of them produces most of the
inflated savings claims we discuss in \cref{sec:componentgap}.

\subsection{Formal cost model}\label{sec:costmodel}

Let a run consume $T_{\mathrm{unc}}$ uncached input tokens,
$T_{\mathrm{cw}}$ cache-creation tokens, $T_{\mathrm{cr}}$ cache-read tokens,
and $T_{\mathrm{out}}$ generated tokens, on a model with input price
$p_{\mathrm{in}}$ and output price $p_{\mathrm{out}}$ (USD per $10^6$
tokens). The billed cost recorded per run is the provider's
\texttt{total\_cost\_usd}; our reconstruction is
\begin{equation}\label{eq:decomp}
\widehat{C} \;=\; \frac{1}{10^6}\Bigl(
T_{\mathrm{unc}}\,p_{\mathrm{in}}
\;+\; T_{\mathrm{cw}}\,\mu_{\mathrm{w}}\,p_{\mathrm{in}}
\;+\; T_{\mathrm{cr}}\,\mu_{\mathrm{r}}\,p_{\mathrm{in}}
\;+\; T_{\mathrm{out}}\,p_{\mathrm{out}}\Bigr),
\end{equation}
with cache multipliers $\mu_{\mathrm{w}}$ (write) and $\mu_{\mathrm{r}}$
(read). Calibration against the billed amounts (\cref{sec:calibration})
selects the provider's published standard prices---Haiku~4.5 at
published list prices per million
input/output tokens---with $\mu_{\mathrm{r}}=0.1$ and
$\mu_{\mathrm{w}}=1.25$ (the five-minute cache TTL)~\citep{anthropic2026pricing,anthropic2026caching}.

Cost per successful execution (CPS) for an arm $a$ over runs $R_a$ with
success set $S_a$ (its successful executions) is
\begin{equation}\label{eq:cps}
\mathrm{CPS}_a \;=\; \frac{\sum_{r \in R_a} C_r}{\lvert S_a\rvert}.
\end{equation}

For a reduction layer $j$, we distinguish its \emph{addressable share}
$A_j$---the fraction of $\widehat{C}$ attributable to context that layer $j$
is able to modify (for a tool-output compressor, the delivered tool text and
its downstream cache traffic)---from its \emph{realized saving}, the paired
change in billed cost,
\begin{equation}\label{eq:realized}
\Delta C_j \;=\; \mathbb{E}_{\text{tasks}}\bigl[\,\overline{C}^{\,(j)}_t -
\overline{C}^{\,(\mathrm{base})}_t\,\bigr],
\end{equation}
estimated over paired blocks (\cref{sec:method}). The two differ by a
\emph{trajectory term}: added or removed turns change every component of
\Cref{eq:decomp} because each turn re-transmits the context prefix. Writing
$\Delta\mathrm{turns}_j$ for the paired change in trajectory length, the
\emph{success-adjusted net effect} of a layer is the paired change in
$\mathrm{CPS}$, which can be negative (a saving) only if per-turn savings
exceed the combined trajectory and success penalties. This is the quantity
we use as the primary decision metric.

\paragraph{Marginal cost of a delivered tool token.} A simplified marginal
model estimates the addressable share. A delivered tool-output token
inserted into the prompt at turn $t$ incurs (per $10^6$ tokens)
\begin{equation}\label{eq:marginal}
C_{\text{tool-token},t} \;=\; \mu_{\mathrm{w}}\,p_{\mathrm{in}}
\;+\; N_{\text{future reads},t}\;\mu_{\mathrm{r}}\,p_{\mathrm{in}},
\end{equation}
where $\mu_{\mathrm{w}}=1.25$ is the observed five-minute cache-write tier,
$\mu_{\mathrm{r}}=0.1$, and $N_{\text{future reads},t}$ is the number of
later model calls that reuse the cached prefix containing the token. Its
first cache creation is priced \emph{above} ordinary input; later reuse is
discounted. An early token is therefore more addressable than a late one:
with the authoritative campaign's mean trajectory of $\approx$4.5 assistant
turns, a token delivered on the first turn is re-read on $\approx$3.5
subsequent calls and costs $\approx(1.25+0.35)\,p_{\mathrm{in}} =
1.60\,p_{\mathrm{in}}$, versus $1.25\,p_{\mathrm{in}}$ for a final-turn
token---and removing an early token also removes its future reads. A removed
token can additionally alter trajectory length itself, so addressability
depends on turn position and future reuse, not on token count alone. The
retained telemetry does not record, per token, the turn of entry and the
number of later calls that reused it, so we present \cref{eq:marginal} as a
bounded conceptual model; exact per-token addressability measurement is
future work.

\subsection{Empirical cost composition}\label{sec:stackpreview}

\begin{figure}[t]
\centering
\includegraphics[width=0.86\linewidth]{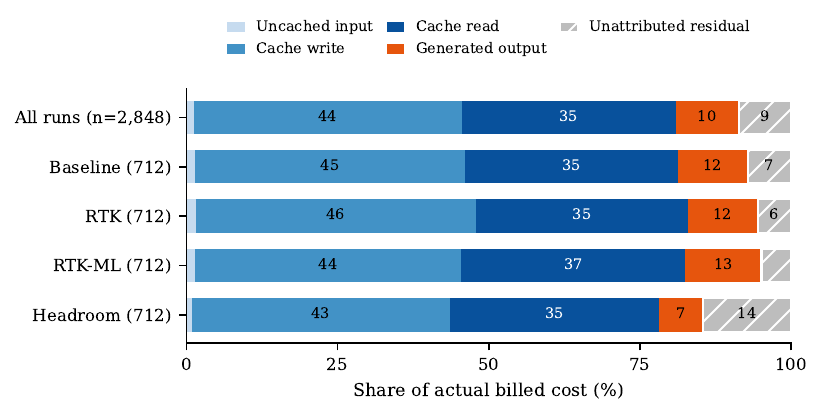}
\caption{Cost decomposition as shares of \emph{actual billed} cost
(n=2,848 analyzed runs; component prices as in \cref{eq:decomp}). The
hatched fifth segment is the unattributed billed-cost residual (billed
minus reconstructed). Numbers inside segments are percentages;
run-bootstrap 95\% CIs for every segment are in \cref{tab:coststack}.
Normalized four-component shares appear in Appendix~\ref{app:normshares}.}
\label{fig:coststack}
\end{figure}

\Cref{fig:coststack} and \cref{tab:coststack} summarize the descriptive result: in the evaluated workloads, generated output is a small
minority of cost, and cache creation plus cache reads account for
approximately 87\% of the reconstructed four-component cost---about 80\%
of the actual bill, with a further 8.7\% of the bill left unattributed by
the reconstruction (shown explicitly as a fifth category). These are
empirical averages from the coding-agent workloads evaluated in this study,
not industry averages; denominators, uncertainty, and caveats are given in
\cref{sec:composition}.

\Cref{tab:componentshares} states the same constraint from the
perspective of a user-side optimization layer, independently of the treatment
comparisons: \emph{which} cost-bearing components such a layer can modify. When the benchmark bill is attributed to the agent component
that generated each measured or inferred contribution, the two surfaces a
user-side layer cannot reach---the harness base (system prompt and tool
definitions) and hidden model thinking with its unattributed replay
residual---together carry $\approx$94.0\% of cost. Every component a layer \emph{can} modify
(tool outputs, tool-call arguments, retrieved file content, injected
context, conversation history) sums to $\approx$6.0\%. That 6\% is the absolute bound under full deletion of every accessible token; since a run must retain sufficient information to succeed, the estimated upper bound for visible-token compression is $\approx$5\% of cost under the measured decomposition. The arms evaluated in this study primarily modified tool outputs, i.e.\ $\approx$3.3\% of cost. This table provides the cost context for the end-to-end results: even complete removal of the accessible component would leave most billed cost attributable to components the layer does not control.

\begin{table}[t]
\centering
\caption{Cost attribution by agent component (benchmark corpus:
1{,}212 end-to-end child sessions, 11{,}773 API requests, reconstructed
from stored transcripts under a calibrated tokenizer; usage fields exact,
component attribution measured, hidden-thinking and cache attribution
inferred). ``Accessible'' marks surfaces a user-side layer can modify.
Shares are of the corpus totals; columns sum to 100\% within rounding.}
\label{tab:componentshares}
\small
\setlength{\tabcolsep}{4.5pt}
\begin{tabular}{lcccl}
\toprule
Component & \% context & \% cache reads & \% cost & Accessible? \\
\midrule
Harness base (system prompt + tool defs) & 71.6 & 92.0 & 74.7 & No \\
Hidden thinking + unattributed residual & 21.8 & 0.0 & 19.4 & Not directly \\
Tool outputs (Bash, tests, Git, edits) & 3.5 & 4.4 & 3.3 & Yes (primary surface) \\
Tool-call arguments (model-emitted JSON) & 1.6 & 1.9 & 1.4 & Yes (untested alone) \\
Retrieved files (\texttt{Read}, web) & 0.9 & 1.0 & 0.8 & Yes \\
Conversation history (user + assistant) & 0.5 & 0.7 & 0.5 & Yes \\
\midrule
Not directly modifiable total & 93.4 & 92.0 & \textbf{94.0} & \\
Accessible total & 6.5 & 8.0 & \textbf{6.0} & \\
\bottomrule
\end{tabular}
\end{table}

\begin{figure}[t]
\centering
\includegraphics[width=0.78\linewidth]{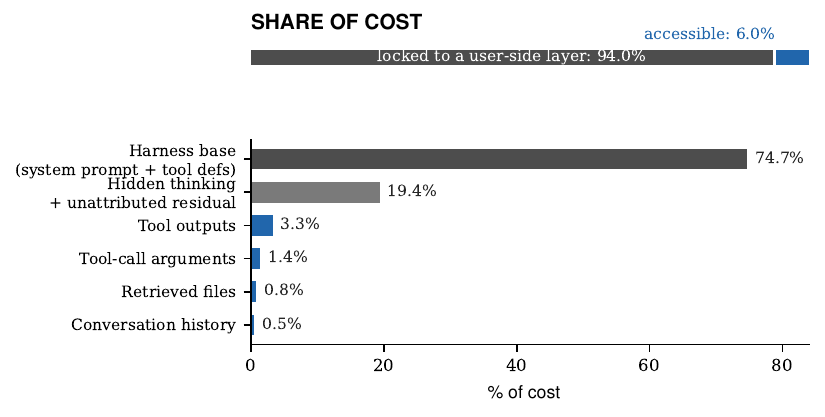}
\caption{Cost attribution by agent component
(\cref{tab:componentshares} as a graphic; benchmark corpus). Gray
components are not directly modifiable by a user-side optimization layer (94.0\% of
cost); blue components are accessible (6.0\%). The estimated upper bound for
visible-token compression is $\approx$5\% of cost.}
\label{fig:componentshares}
\end{figure}

\begin{figure}[t]
\centering
\includegraphics[width=0.78\linewidth]{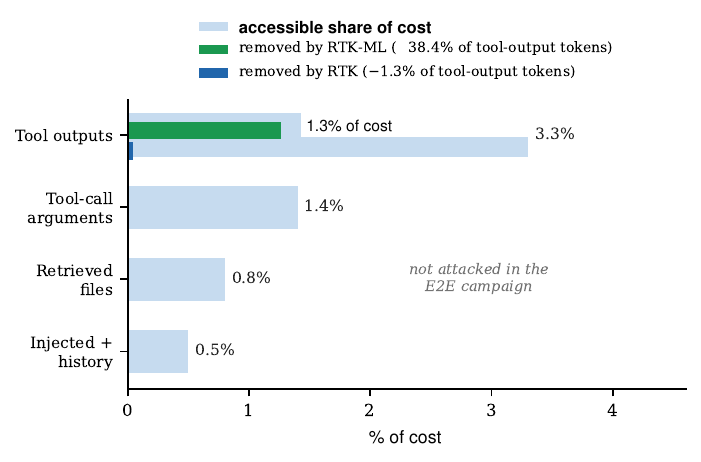}
\caption{Realized reduction per accessible component. Light bars: each
component's share of cost. Overlaid bars: the share actually
removed by the evaluated hook layers, applying the hook-side ledger's
raw-vs-delivered tool-output reduction ($-38.4\%$ RTK-ML, $-1.3\%$ RTK;
\cref{sec:decoupling}) to the tool-output share. Even the campaign's
largest observed reduction reaches $\approx$1.3\% of cost---and
that arm's paired billed cost was \emph{higher} than baseline
(\cref{tab:aggregate}).}
\label{fig:componentsavings}
\end{figure}

\Cref{fig:componentshares} renders the same attribution graphically, and
\cref{fig:componentsavings} shows what the
evaluated compressors actually removed on each accessible surface: the
largest observed tool-output reduction in the campaign---RTK-ML's $-38.4\%$ of raw
tool-output tokens (\cref{sec:decoupling})---amounts to $\approx$1.3\% of
total cost, the shipped RTK's $-1.3\%$ to $\approx$0.04\%, and no evaluated layer modified the remaining accessible components.

\section{Optimization Surfaces and Architectures}\label{sec:architectures}

Context-reduction systems differ in \emph{where} they act on the agent loop,
what they can see, and whether their transformations are reversible.
\Cref{tab:archmatrix} summarizes the systems studied or audited here;
evidence levels refer to the layered taxonomy of \cref{sec:componentgap}
(L1--L8).

\begin{table}[t]
\centering
\caption{System architecture and evidence matrix. ``Billed E2E'' means paired
end-to-end campaigns with provider-billed cost and deterministic task
scoring exist in the retained artifacts. Evidence levels: L1 compression
ratio, L2 retrieval/preservation quality, L3 model-visible context change,
L4 production-path activation, L5 task success, L6 billed cost, L7 cost per
successful execution, L8 trajectory/failure analysis.}
\label{tab:archmatrix}
\small
\setlength{\tabcolsep}{4pt}
\resizebox{\textwidth}{!}{%
\begin{tabular}{llllll}
\toprule
System & Surface & Visibility & Reversible & Prod.\ wiring & Evidence \\
\midrule
Baseline Claude Code & --- (reference) & full harness & --- & shipped & L4--L8 \\
RTK & PreToolUse hook; CLI proxy & raw+delivered tool I/O & rules only & shipped & L1--L8 \\
\Ninegate{} & hook + flag-gated gate stack & raw+delivered tool I/O & byte-equiv.\ on covered disabled paths & benchmark build & L1--L8 \\
Headroom v0.27.0 & API-boundary proxy & opaque (black box) & unknown & shipped (OSS) & L5--L8 \\
Vendor CLI compressor & CLI output compressor & external; not run here & n/a & external & vendor claims only \\
\texttt{ml\_lexical} engine & query-aware rank/pack engine & component harness & fail-open & research code & L1--L2 \\
Caveman & agent-prose compressor & component harness & preservation contract & not registered & L1--L2 \\
Lexical ranking & candidate re-ranker (gate 4) & candidates & reorder only & flag-gated & L2, L4 \\
Embedding ranking & BGE sidecar re-ranker (gate 5) & candidates & reorder only & flag-gated & L4 (active, not decisive) \\
Structural (ast-grep) & AST-anchored retrieval (gate 8) & focus hints & additive only & flag-gated & L2, L4 \\
Union/hybrid retrieval & lexical$\cup$embedding & candidates & reorder only & component harness & L2 \\
\bottomrule
\end{tabular}
}
\end{table}

\paragraph{Hook-based compression (RTK).} RTK installs a
\texttt{PreToolUse} hook that rewrites eligible shell commands to route
through a CLI proxy; the proxy applies YAML-configured, strategy-based
compression to the command's output before it reaches the model. RTK is
the shipped deterministic layer. The \emph{\ninegate} architecture
adds nine independently flag-gated capabilities on top: (1) query
propagation from the user prompt into the proxied command; (2) retrieval
routing of search-type commands (\texttt{grep}, \texttt{rg}, \texttt{find},
\texttt{ls}, \ldots) into a grouped, ranked \emph{search-group}
representation; (3) query-aware deterministic re-ranking; (4) a lexical
scoring leg; (5) an embedding scoring leg backed by a loopback BGE
sidecar~\citep{xiao2023bge}; (6) a task-family router that classifies the
context; (7) a preserve gate that bypasses compression entirely for
high-recall families; (8) structural retrieval using
ast-grep~\citep{astgrep2026} to inject focus-file hints; and (9) a
session-scoped focus cache. All nine are off by default, and the disabled
paths covered by the dedicated equivalence tests were byte-equivalent; a
safety router
(default-on) prevents semantic stages from touching sensitive tool output,
and a stdout-consumer guard (\cref{sec:failures}) suppresses ranking when
the command's stdout feeds another program.

\paragraph{API-boundary proxying (Headroom).} Headroom~\citep{headroom2026}
interposes on the provider API endpoint via \texttt{ANTHROPIC\_BASE\_URL}
and optimizes payloads in flight. Its observed configuration in these
campaigns was token-mode optimization with caching enabled and code-aware
mode disabled. Because it is a black box to our harness, we can measure its
billed effects and trajectories but cannot attribute them to internal
mechanisms; we therefore avoid causal claims about \emph{why} it behaves as
it does.

\paragraph{Deterministic CLI compression (external vendor tool).} One
audited system is a commercially distributed deterministic command-output
compressor from an unaffiliated vendor. The retained artifacts contain
only that vendor's own per-category claimed savings (75--80\% for
\texttt{git diff}, \texttt{git log}, \texttt{ls}, \texttt{grep}),
embedded as reference literals in a benchmark harness. No same-harness,
paired, actually-billed end-to-end run of it exists in the artifacts, so
it appears in this paper only as a related architectural approach.

\paragraph{Query-aware research engine (\texttt{ml\_lexical}).} The
\texttt{ml\_lexical} engine is an implemented research architecture combining
chunking, lexical and cosine ranking, packing with protected regions, and a
BGE embedding sidecar. Its evidence is component-level retrieval quality
only (\cref{sec:benchmarks}); it has no end-to-end task-success or billed
cost evidence.

\paragraph{Agent-prose compression (Caveman).} Caveman compresses
LLM-generated prose (never tool output) under an explicit preservation
contract (paths, line numbers, test names, error types, metrics, and code
blocks are kept verbatim). It is not registered in any production strategy
table; its evidence is a seven-fixture component study
(\cref{sec:benchmarks}).

\section{Benchmarks and Datasets}\label{sec:benchmarks}

\Cref{tab:benchmatrix} maps every benchmark used in the underlying research
program to what it does and does not measure. Two properties matter most:
whether the benchmark observes \emph{end-to-end task success}, and whether
its costs are \emph{actually billed} rather than reconstructed.

\begin{table}[t]
\centering
\caption{Benchmark capability matrix. ``Billing'' distinguishes: provider-returned usage with execution-result
cost cross-checked against the pricing schedule (billed+usage); cost
reconstructed from provider-returned token counts (tokens); billed cost
recovered from frozen reports, ledgers, or traces (billed, with the source
named); and length-based estimates (est.).}
\label{tab:benchmatrix}
\small
\setlength{\tabcolsep}{3.5pt}
\resizebox{\textwidth}{!}{%
\begin{tabular}{lllllp{4.1cm}}
\toprule
Benchmark & $n$ used & Gold / success metric & Level & Billing & Key limitation \\
\midrule
ContextBench~\citep{contextbench2026} & 36 sweep runs / 7,510 rows; 50 (policy study) & gold-span token survival & component & tokens & no task success; dataset revision unrecorded \\
LoCoEval~\citep{locoeval2026} & 50 (preservation; retained) & reference-answer survival & component & est.\ (len/4) & judge unavailable; preservation proxy only \\
InterCode-proxy~\citep{yang2023intercode} & 24 obs.\ / 240 rows (12 MBPP~\citep{austin2021mbpp} $\times$ pass/bug) & marker recall $\geq$95\% & component & tokens & interactive task success blocked (no container runtime) \\
Structural retrieval & 8 tasks / 38 queries & test-grounded def-site gold & component & tokens & agent success not measured \\
Caveman fixtures & 7 synthetic & critical-marker drops & component & tokens & synthetic; not production traffic \\
ContextBench-Go grounded E2E & 29 rows $\times$ 4 arms & patch applies + tests resolve & E2E (single-shot) & billed (usage) & one model; single-shot shape (\cref{sec:grounded}) \\
SWE-bench\_Pro Go agent smoke & 4$\times$4 + 18 trials & patch apply / resolve & E2E (single-shot) & billed (report-derived) & 0 applies context-only; smoke scale \\
IC-SWE flask pilot & 3 tasks $\times$ 4 modes & FAIL\_TO\_PASS flips & E2E (single-shot) & billed (trace-derived) & n=3, single repo; labeled pilot \\
Comprehensive E2E pilot & 21 tasks $\times$ 4 systems & pytest exit + diff constraints & E2E & billed (report-derived) & single model, 1 repetition \\
Large-session campaign & 529 sessions, 3 tasks & pytest exit & E2E & billed (ledger) & different RTK binary lineage \\
Authoritative campaign & 103 tasks, 2,908 runs & deterministic judges & E2E & billed+usage (cross-checked) & Claude-Code-specific harness \\
Codex Headroom replication & 40 tasks $\times$ 2 arms (+12$\times$2 pilot) & deterministic graders & E2E & tokens (reconstructed) & different agent (Codex CLI); transcribed from frozen report (\cref{sec:codex}) \\
\bottomrule
\end{tabular}
}
\end{table}

\paragraph{Limits of gold-context preservation as an agent metric.} Component
benchmarks score a compressor against human- or test-annotated \emph{gold}
context. Four mechanisms can cause gold-context preservation metrics to diverge from
agent outcomes. (i)~Agents can \emph{recover}: if a needed line is removed,
the model can search again or re-read the file, converting a preservation
failure into a trajectory cost rather than a task failure. (ii)~Agents can
\emph{succeed without the gold}: prior knowledge or inference over remaining
context is often sufficient. (iii)~Agents can \emph{fail with the gold}:
receiving the right lines does not guarantee correct edits. (iv)~Retrieval
is one stage of a multi-stage trajectory; a single extra
diagnosis-and-retry loop re-transmits the full context prefix and can
eliminate the nominal savings of a much larger compression
(\cref{sec:phases}). Long-context research documents related gaps between
context presence and context use~\citep{liu2024lost,bai2024longbench}.

\paragraph{Suite-level benchmarks and created corpora.} Beyond the
benchmarks above, the program built and ran a family of suite-level
corpora: a 159-case command-output compression corpus (\texttt{cmdcompress})
with 34 paired live-API cases, privacy-redaction and intent-classification
suites (\texttt{privacybench}, \texttt{observer\_intent}), retrieval-mode
and output-compression exercises, and a path-coverage suite proving which
changed code each arm actually executes. It also \emph{created} the
synthetic task fixtures the end-to-end campaigns run on: seven scripted
repository tasks for the pilot harness, fourteen fixture suites (343 files,
including 150/400-test pass/fail suites and bulk-output generators) for the
comprehensive campaign, three synthetic large repositories (911 files) for
the long-session marathons, seeded synthetic monorepos with committed
ground truth for Stages 1--2 and the authoritative campaign, the 7
Caveman prose fixtures, and the 24-observation captured pytest corpus.
Appendix~\ref{app:corpora} inventories all of them with sizes and paths.

\paragraph{SWE-bench lineage.} The grounded-completion program
(\cref{sec:grounded}) draws its tasks from SWE-bench-derived corpora: the
ContextBench-Go rows are Multi-SWE-bench Go instances
\citep{zan2025multiswebench} exposed through the ContextBench loader, the
agent smoke ran on the \texttt{Contextbench/SWE-bench\_Pro} dataset
\citep{swebenchpro2026}, and the IC-SWE pilot uses SWE-bench Python
repositories at pinned base commits \citep{jimenez2024swebench}. Task
success is scored by whether the row's FAIL\_TO\_PASS tests flip to passing
after the agent's patch applies.

\paragraph{Snapshots.} The authoritative campaign is fully frozen (task
manifests, six pinned repository SHAs plus a seeded synthetic monorepo,
sha256-pinned binaries, gzipped transcripts). The component-benchmark side is
weaker: the ContextBench dataset revision hash was not recorded (row
identifiers are retained), and parts of the LoCoEval artifacts were written
to volatile temporary directories. Appendix~\ref{app:snapshots} lists the
exact snapshot state per dataset.

\section{Experimental Methodology}\label{sec:method}

\subsection{Campaign hierarchy}\label{sec:campaigns}

The underlying research program produced several campaign generations
(\cref{tab:campaigns}). The program's evidence classes are reconciled
exactly in Appendix~\ref{app:reconciliation}: 5{,}123 executions recorded
in append-only runtime cost ledgers (guarded
re-executions such as the Stage-1 post-guard rerun and the two
separately-billed long-session sibling campaigns counted at their true
cost); 8 costsmoke executions whose count derives from the suite
results file rather than a per-run ledger; and 362 billed single-shot
trials whose reported or trace-reconstructed spend is itemized in
\cref{tab:benchprogram} (analyzed in \cref{sec:grounded}). The
combined measured program therefore comprises 5{,}493 billed executions,
8{,}263 zero-inference-cost
component evaluations, and a free deterministic gate-activation proof
(Stage~0). Later campaigns supersede earlier ones where they overlap.
\Cref{tab:campaigns} lists them with their role in this paper. All
quantitative end-to-end claims in
\Cref{sec:composition,sec:results,sec:heterogeneity,sec:phases} come from
the \emph{authoritative campaign} unless explicitly labeled otherwise; the
long-session campaign is reported separately because it used an earlier
experimental configuration and must not be pooled with the authoritative
campaign; earlier pilots are cited only
for provenance and for the failure analysis of \cref{sec:failures}.

\begin{table}[t]
\centering
\caption{Campaign inventory and role, entire measurement program.
Execution counts are totaled from each campaign's append-only runtime cost
ledger (every billed execution counted once, including guarded
re-executions); analysis populations de-duplicate rescored rows separately.
The authoritative campaign comprises a base phase and a pre-specified
expansion (new hash-frozen holdout) analyzed jointly. Stage 0 is a free
deterministic proof and is excluded from the paid total.}
\label{tab:campaigns}
\small
\setlength{\tabcolsep}{3.6pt}
\resizebox{\textwidth}{!}{%
\begin{tabular}{llllp{4.0cm}}
\toprule
Campaign & Runs & Models & Systems & Role in this paper \\
\midrule
E2E smoke pilot & 32 & Haiku 4.5 & 4 & provenance only \\
Small comparison (costsmoke) & 8\textsuperscript{d} & Opus 4.8 & 4 & suite-level; superseded on compression axis \\
Comprehensive E2E pilot & 96 & Haiku 4.5 & 4(+4 modes) & Headroom 18/21 discussion (\cref{sec:headroom1821}) \\
Expanded statistical (API) & 112 & Opus 4.8 & 4 & component-level; compression at scale \\
Large-session (2-system gen.) & 133 & Opus 4.8 & 2 & earlier generation, superseded by 4-system \\
Large-session campaign & 529 & Opus 4.8 & 4 & long-session evidence; separate binary lineage \\
Large-session attribution & 60 & Opus 4.8 & 4 & attribution follow-up (same lineage) \\
Stage 1 (pre-guard + rerun) & 911 & Haiku 4.5 & 5 & pipeline-incompatibility analysis + post-guard rerun \\
Stage 2 (paid, post-guard) & 342 & Haiku, Sonnet & 4 & guard validation; model-reversal observation \\
\textbf{Authoritative campaign} & \textbf{2,908} & \textbf{Haiku, Sonnet, Opus} & \textbf{4} & \textbf{all primary results} \\
\midrule
Single-shot grounding program & 362 & Sonnet 4.6 & (see \cref{tab:benchprogram}) & anchor alteration + grounding study (\cref{sec:grounded}) \\
\midrule
\textbf{Paid total} & \textbf{5{,}493} & --- & --- & Stage 0 adds 4 free checks; \textsuperscript{d}\,= count derived from suite results, no per-run ledger \\
\bottomrule
\end{tabular}
}
\end{table}

\begin{table}[t]
\centering
\caption{The single-shot grounded-completion and agent-smoke program
(companion bench harness; claude-sonnet-4-6; billed
\texttt{anthropic\_api\_usage}). Trials are (row $\times$ arm) agent calls;
$\sim$ marks report-approximate or trace-derived spend.}
\label{tab:benchprogram}
\small
\setlength{\tabcolsep}{4pt}
\resizebox{\textwidth}{!}{%
\begin{tabular}{lcp{8.2cm}}
\toprule
Experiment (single-shot, Sonnet 4.6) & Trials & Result \\
\midrule
SWE-bench\_Pro agent smoke & 16 & oracle 2/4 resolved; Claude patches generated, 0 applied \\
ContextBench-Go C3/D3/D3b/E investigation & 84 & Phase E (n=40 x 2 arms): raw 2/40 vs compressed 1/40 resolved; apply 27/40 vs 15/40 \\
SWE-bench\_Pro Go probe+confirm & 18 & 0/18 applied, 0 solved (context-only agent, no verbatim anchors) \\
G2 grounded 4-arm (29 oracle rows) & 116 & grounded raw 5/29 solved; 3 rows solved only by grounded raw, 0 only by grounded compressed (exact paired p=0.25) \\
Family-routing E2E validation & 116 & preserve 83\% vs aggressive 48\% apply; grounded preserve 6 solves \\
IC-SWE flask agent pilot & 12 & 0/3 all modes; hybrid-lexical 34.7\% savings, 67\% patch-apply \\
\midrule
\textbf{Program subtotal} & \textbf{362} & plus 8,263 zero-inference-cost component evaluation rows \\
\bottomrule
\end{tabular}
}
\end{table}

\subsection{Design of the authoritative campaign}\label{sec:design}

\paragraph{Arms.} (1)~\emph{Baseline}: Claude Code 2.1.201 with no layers.
(2)~\emph{RTK v0.44.1}: the open-source RTK hook layer used for all
RTK experiments in this paper. (3)~\emph{\Ninegate}: the flag-gated architecture at
the guard commit, all nine gates enabled, safety router in enforce mode,
BGE embedding sidecar health-checked before every batch.
(4)~\emph{Headroom v0.27.0} via \texttt{ANTHROPIC\_BASE\_URL}. Binary
sha256 hashes and exact provenance are in Appendix~\ref{app:provenance}.
All RTK rows in the paper refer to v0.44.1. The post-hoc Opus-medium
replication uses the same RTK version and is reported separately only
because it was run after the main campaign under Claude Code 2.1.220
(\cref{app:upstream}).

\paragraph{Tasks and repositories.} 103 analyzed tasks across 24 defined
families (23 with analyzed tasks; the one family without an analyzed task
contained only the excluded defective task)
(\cref{sec:heterogeneity}) over 7 repositories: six pinned open-source
repositories (click, cobra, express, flask, gin, requests) and a seeded
synthetic monorepo. Tasks are frozen in pre-specified, hash-frozen manifests with
per-task judges (deterministic answer/regex/test checks), allowed/forbidden
file constraints, timeouts, and design attributes (difficulty, retrieval
mode, session band, output band).

\paragraph{Pairing and randomization.} The unit of execution is a
\emph{block}: one task $\times$ model $\times$ effort $\times$ repetition,
in which all four arms run from identical fresh working copies in
randomized order. Runs execute in an isolated per-run \texttt{\$HOME} with a
scrubbed environment (all \texttt{CLAUDE*}/\texttt{ANTHROPIC*} and
compression-layer variables removed except the API key), fixed harness
settings (project-only setting sources so no user hooks load), per-run
budget caps, and a global spend cap enforced under a lock.

\paragraph{Models and effort.} Claude Haiku 4.5, Claude Sonnet 5, and
Claude Opus 4.8, at effort levels low/medium/high/xhigh/max for Haiku and
low/high for Sonnet and Opus (unmeasured cells are reported as such,
never imputed; \cref{sec:heterogeneity}).

\paragraph{Measurement.} Cost is the provider-billed
\texttt{total\_cost\_usd} from the harness JSON output
(field \path{cost_source} equal to \path{actual_billed} on every
analyzed run). For every execution, \texttt{total\_cost\_usd} was obtained
from the Claude Code/API execution result and independently cross-checked
against the provider-returned usage fields and the applicable Anthropic
pricing schedule; it was not estimated from text length or local token
heuristics. Usage tokens come from the provider usage object; success is scored by the frozen
deterministic judges; gate activation is read from per-run gate traces;
raw-vs-delivered tool tokens come from the hook-side ledger (RTK
arms only), counted with an embedded local BPE tokenizer (estimates, not
provider token counts). Every transcript is retained gzipped, enabling the turn-level
ledgers and phase analysis of \cref{sec:phases}.

\paragraph{Pre-specification and split roles.} The campaign design, arms,
tasks, judges, and analysis plan were frozen in hash-pinned manifests and
pre-specification documents before execution (the expansion added a new
frozen holdout before its spend). There was no public, independently
timestamped preregistration; we therefore use ``pre-specified'' throughout
rather than ``pre-registered.'' Findings are labeled by provenance:
development/base-split observations (which also drove the guard fix and
judge repairs), frozen-holdout confirmations (\cref{tab:holdout}), pooled
estimates, and exploratory strata.

\paragraph{Exclusions and integrity.} Analysis rows are de-duplicated
keep-last per run id (rescored rows are appended, never overwritten),
restricted to completed non-infrastructure-failure runs, and exclude one
manifest-flagged defective task and the calibration split: 2,848 analyzed
runs out of 2,908 executed. Five judge-defect classes
found before the holdout were fixed and rescored symmetrically across all
arms from stored outputs, with the full append-only row history retained.

\subsection{Prompt-cache carryover between arms (threat to validity)}
\label{sec:carryover}

Because the measured outcome is directly affected by provider-side prompt
caching, within-block cache carryover is a first-order threat: all four
arms of a block run the same task from identical working copies, per-run
\texttt{\$HOME} isolation does not isolate the provider's cache, and the
median gap between consecutive runs in a block is 10.8\,s (p90 28.6\,s;
100\% of the 2,136 consecutive pairs fall inside the five-minute cache
TTL). A TTL-separated re-analysis is therefore impossible on this data ---
carryover cannot be ruled out by temporal separation. Three retained-data
checks bound the concern (\cref{tab:order}; full numbers in the
reproducibility package). First, the design is exactly balanced at the margins: each arm ran 712
times in total and each block position contains exactly 712 runs
(arm-by-position cell counts vary 147--221 under randomization, which is
not stratified by position). Second, the primary paired
deltas do not depend on relative order: for every arm, the delta measured
when the arm ran \emph{before} baseline is statistically indistinguishable
from the delta when it ran \emph{after} (all order-difference CIs cross
zero). Third, later block positions show no cache-write deflation---mean
cache-creation tokens at positions 1--3 are 3.7\% \emph{higher} than at
position 0, the opposite of what shared-prefix reuse would produce, and
mean cost varies non-monotonically by position (a 9\% spread).
Randomized order limits systematic bias by construction, and these checks
detect no order signature in cost, cache traffic, or success; they do not
prove carryover is absent, and we carry the point as a limitation.

\begin{table}[t]
\centering
\caption{Order-sensitivity check for the main 2,848-run campaign:
paired per-task cost delta vs.\ baseline, split by whether the arm executed
before or after baseline within its randomized block. All order-difference
CIs cross zero. The post-hoc RTK v0.44.1 Opus-medium replication is not pooled into this
table because it is a different campaign (\cref{app:upstream}).}
\label{tab:order}
\small
\begin{tabular}{lcccc}
\toprule
Arm & $\Delta$\% (arm first) & $\Delta$\% (arm after) & Order difference (\%) [95\% CI] & $n_T$ \\
\midrule
RTK v0.44.1 & -4.41 & -1.57 & -2.84 [-9.54, +3.75] & 101/99 \\
RTK-ML & +11.53 & +4.04 & +7.49 [-2.32, +18.01] & 99/101 \\
Headroom v0.27.0 & +47.31 & +49.69 & -2.40 [-17.57, +12.02] & 102/99 \\
\bottomrule
\end{tabular}
\end{table}

\subsection{Statistical procedure}\label{sec:stats}

The task is the unit of inference for comparative arm effects.
Descriptive cost-composition and reconstruction intervals use run-level
resampling and are not interpreted as task-level treatment-effect
inference. For any comparative metric we form paired
within-block differences (arm minus baseline), average them within task,
and bootstrap over tasks (10{,}000 resamples, fixed seed) for 95\%
confidence intervals; the expansion analysis additionally reports
repository-clustered bootstrap intervals and leave-one-repository-out
sensitivity for the same contrasts. Resampling units by analysis: arm
comparisons use the task-clustered bootstrap; cost-composition shares and
the reconstruction residual use a run-level bootstrap (descriptive only;
run-level resampling does not account for repeated-task dependence);
repository sensitivity uses a repository-clustered bootstrap; grounded
paired outcomes use an exact paired analysis; exploratory task-family
cells are descriptive or task-bootstrap as flagged in each table. Repeated runs of the same task are
never treated as independent: intraclass correlation of cost across
repetitions was 0.37--0.55 depending on arm, giving Kish effective sample
sizes of roughly 38--45 tasks per arm despite 712 runs per arm.
Family-level results are exploratory screening: the underlying campaign
screened families by whether task-clustered bootstrap intervals excluded
zero, but no per-contrast $p$-values were computed, so no
false-discovery-rate control is claimed; we report family-level effect
sizes with confidence intervals and label them exploratory. Cells with fewer
than six tasks are flagged unstable and reported as exploratory. We report
effect sizes with confidence intervals and avoid binary significance
language where power is insufficient. Failed tasks are never dropped from
cost aggregates, and lower cost caused by early failure is not treated as
efficiency: cost per successful execution (\cref{eq:cps}) keeps failed runs in
the numerator.

\section{Aggregate Cost Composition}\label{sec:composition}

\subsection{Calibration of the decomposition}\label{sec:calibration}

Applying \cref{eq:decomp} with the provider's published standard prices and
the five-minute cache-write multiplier ($\mu_{\mathrm{w}}{=}1.25$,
$\mu_{\mathrm{r}}{=}0.1$) reproduces the billed cost of individual runs with
median per-run residuals of $+0.9\%$ (Haiku, $n{=}1{,}748$), $+0.0\%$
(Sonnet, $n{=}948$), and $+0.2\%$ (Opus, $n{=}152$). Alternative
configurations are rejected by the same calibration: the one-hour
cache multiplier leaves $-23$\% to $-33$\% median residuals, and Sonnet~5
introductory pricing leaves $+33\%$ unexplained---so the campaign was
billed at standard prices under the five-minute cache tier.

The median residual and the aggregate residual answer different questions,
and the distinction matters for all reported results in this paper. The
four-component model is accurate for most \emph{individual} runs (the
median residuals above), but the \emph{dollar-weighted aggregate} residual
is $+8.7\%$ of total billed spend (95\% CI \ci{7.2}{10.1}; by arm:
baseline $+7.1\%$, RTK $+5.5\%$, \ninegate{} $+4.9\%$, Headroom
$+14.5\%$)---an economically material share that cannot be explained by
the small median. A post-hoc per-model analysis (script
\path{scripts/extract_thinking_addressability.py}; retained run records
only) localizes it. On Sonnet~5 the four components account for the bill
at both measured efforts (residual $+0.2\%$), so its aggregate
\texttt{output\_tokens} appear to include thinking charges. On Haiku~4.5
the residual rises monotonically with the thinking-effort setting---from
11.0\% to 18.3\% of the cell's bill per run (rising monotonically;
implied $+753$ to $+1{,}283$ tokens at the output price)---while measured
output ($\approx$780/run), cache volumes, and turn counts stay flat.
Cache-tier and turn-count explanations are ruled out by those flat
covariates, and secondary-model auxiliary calls contribute only a smaller,
flat component (residual 5.2\% for one-model vs.\ 7.2\% for two-model
runs): the effort-scaling component is therefore
\emph{thinking-consistent}, and on this model thinking charges sit inside
the residual rather than the measured output component. Opus~4.8 cells are
too small to conclude (n=24/90). Provider usage semantics are thus not
uniform across models in this harness---the same aggregate field appears
to include thinking on Sonnet~5 and exclude it on Haiku~4.5. The
reconstruction therefore does \emph{not} account for every billed dollar,
and all reported shares are reported against the actual bill with the
residual shown explicitly; the ``generated output'' component is
\emph{measured} output. The larger Headroom residual is an observation,
not evidence of any particular internal mechanism.

\subsection{Billed cost decomposition}\label{sec:where}

\begin{table}[t]
\centering
\caption{Cost decomposition as shares of \emph{actual billed} cost with
run-bootstrap 95\% CIs, by slice, including the unattributed residual
(billed minus reconstructed). Population: 2,848 analyzed runs from the
main campaign; the RTK row therefore refers to RTK v0.44.1 in the main campaign;
the post-hoc Opus-medium cells are reported separately. Cache write plus cache read is
$\approx$80\% of the bill ($\approx$87\% of the reconstructed
four-component cost). Normalized four-component shares for finer slices are
in Appendix~\ref{app:normshares}.}
\label{tab:coststack}
\small
\setlength{\tabcolsep}{3.2pt}
\resizebox{\textwidth}{!}{%
\begin{tabular}{lccccc}
\toprule
Slice & Uncached & Cache write & Cache read & Output & Unattributed residual \\
\midrule
All runs & 1.3 [1.1, 1.5] & 44.3 [43.2, 45.3] & 35.4 [34.5, 36.3] & 10.4 [10.0, 10.9] & 8.7 [7.2, 10.1] \\
Baseline & 1.4 [1.0, 1.8] & 44.7 [42.9, 46.4] & 35.2 [33.6, 36.6] & 11.7 [10.7, 12.6] & 7.1 [4.6, 10.1] \\
RTK v0.44.1 & 1.5 [1.1, 2.0] & 46.4 [44.7, 48.2] & 35.0 [33.5, 36.5] & 11.5 [10.7, 12.4] & 5.5 [3.0, 8.4] \\
RTK-ML & 1.4 [1.0, 1.8] & 44.0 [42.2, 46.1] & 37.1 [35.2, 38.8] & 12.6 [11.6, 13.6] & 4.9 [2.9, 7.4] \\
Headroom & 0.9 [0.6, 1.1] & 42.8 [40.4, 45.4] & 34.6 [32.6, 36.8] & 7.3 [6.6, 7.9] & 14.5 [11.0, 18.0] \\
\bottomrule
\end{tabular}
}
\end{table}

Across all 2,848 analyzed runs (\cref{tab:coststack},
\cref{fig:coststack}), as shares of the \emph{actual bill}: cache creation
44.3\% (CI \ci{43.2}{45.3}), cache reads 35.4\% \ci{34.5}{36.3}, generated
output 10.4\% \ci{10.0}{10.9}, uncached input 1.3\% \ci{1.1}{1.5}, and
unattributed residual 8.7\% \ci{7.2}{10.1}. Equivalently, cache creation
and cache reads accounted for approximately 87\% of the reconstructed
four-component cost (48.5\% and 38.7\% of $\widehat{C}$) and about 80\% of
the bill itself. The mean run carried $\approx$117k cache-read tokens and
$\approx$12k cache-creation tokens against only $\approx$715 generated
tokens.

This composition, rather than any property of a specific compression layer, is the paper's main explanatory result for RQ1. A system may remove a large fraction of a particular tool output while changing only a small fraction of total billed cost because that output may represent only a small part of the complete prompt-cache and trajectory cost.

Slice-level analysis provides additional detail (normalized four-component shares;
Appendix~\ref{app:normshares}). Failed runs shift spend toward cache reads
(43.5\% vs.\ 38.6\% of $\widehat{C}$ on successes) and generated output
(14.1\% vs.\ 11.3\%)---a pattern consistent with additional re-reading or
recovery behavior. Long
sessions raise the output share from 7.4\% (short) to 18.4\% and nearly
triple cache-read tokens per run. Headroom's profile has the lowest output
share (8.5\%) and the highest cache share of any arm; the tested
configuration was associated with higher cache-side usage; Opus is the only model
with a material uncached-input share (11.3\%). Because cache reads are
billed at $0.1\times$ input price, the \emph{addressable} saving from
removing one delivered tool token is an order of magnitude smaller than
nominal token counts suggest once that token has entered the cached
prefix---and is then re-read on subsequent turns at the discounted rate. Consequently, end-to-end agent cost cannot be inferred from a simple input-versus-output token decomposition (RQ2).

\section{End-to-End Results}\label{sec:results}

\subsection{Aggregate outcomes}\label{sec:aggregate}

\begin{table}[t]
\centering
\caption{Aggregate end-to-end results, authoritative campaign (per arm:
712 runs, 103 tasks). CPS = cost per successful execution
(\cref{eq:cps}). $\Delta$cost is the paired per-task billed-cost
change vs.\ baseline as \% of the baseline mean, task-clustered bootstrap
95\% CI; $\Delta$succ.\ is the paired success change in percentage points.}
\label{tab:aggregate}
\small
\setlength{\tabcolsep}{4pt}
\begin{tabular}{lccccc}
\toprule
System & Runs & Succ.\ (\%) & Turns & $\Delta$cost (\%) & $\Delta$succ.\ (pp) \\
\midrule
Baseline Claude Code & 712 & 96.2 & 4.46 & --- & --- \\
RTK v0.44.1 & 712 & 96.4 & 4.32 & -2.7 [-5.6, -0.1] & -0.2 [-2.9, +1.9] \\
RTK-ML & 712 & 97.5 & 4.69 & +6.8 [+2.8, +11.3] & +0.9 [-0.6, +2.5] \\
Headroom v0.27.0 & 712 & 97.8 & 4.20 & +48.4 [+42.3, +55.0] & +1.4 [-0.2, +3.1] \\
\bottomrule
\end{tabular}
\end{table}

\Cref{tab:aggregate} reports the primary end-to-end comparison. No
success-rate difference was detected within the precision of the campaign
(all paired success CIs cross zero; \ninegate{} $+0.9$\,pp \ci{-0.6}{+2.5},
Headroom $+1.4$\,pp \ci{-0.2}{+3.1}); the study was not designed as a
formal non-inferiority trial, and the high baseline success rate
(96--98\%) limits sensitivity to small performance losses
(\cref{sec:limitations}). Costs, by contrast, separate clearly: RTK v0.44.1 is
slightly lower than baseline ($-2.7\%$, CI \ci{-5.6}{-0.1}); the \ninegate{} build
is a small net cost ($+6.8\%$, CI \ci{+2.8}{+11.3}); and Headroom carries a
large, consistent cost increase ($+48.4\%$, CI \ci{+42.3}{+55.0}). The
expansion analysis confirmed the Headroom and RTK-ML-vs-RTK orderings
under repository-clustered intervals and leave-one-repository-out
sensitivity. \Cref{tab:holdout} separates the frozen-holdout estimates
from the pooled ones: the Headroom cost increase ($+47.3\%$ \ci{+35.2}{+59.2})
and the \ninegate{} cost ($+9.1\%$ \ci{+2.4}{+16.9}) are confirmed on the
holdout alone, whereas RTK v0.44.1's small saving is
a pooled estimate whose holdout-only interval crosses zero ($-2.3\%$
\ci{-7.4}{+2.1})---it should be read as suggestive, not
holdout-confirmed. A post-hoc paired replication extended RTK v0.44.1 to cells the main
campaign did not measure---Opus 4.8 and Opus 5 at medium effort. RTK
v0.44.1 changed billed cost by $-2.9\%$ \ci{-18.8}{+9.3} on Opus~4.8 and
by $+2.5\%$ \ci{-6.1}{+12.0} on Opus~5; pooled across those cells the
change was $-0.1\%$ \ci{-9.3}{+7.8}. The corresponding CPS ratios are
0.933 \ci{0.795}{1.073}, 1.025 \ci{0.941}{1.123}, and 0.979
\ci{0.896}{1.066} pooled (\cref{tab:cps,app:upstream}).

\begin{table}[t]
\centering
\caption{Frozen-holdout-only vs.\ pooled paired billed-cost change vs.\
baseline. Findings first observed during development and confirmed on the
frozen holdout are distinguished from pooled estimates.}
\label{tab:holdout}
\small
\begin{tabular}{lcc}
\toprule
Arm & Frozen holdout only & Pooled (development + holdout) \\
\midrule
RTK v0.44.1 & -2.3\% [-7.4, +2.1] ($n_T$=27) & -2.7\% [-5.6, -0.1] ($n_T$=103) \\
RTK-ML & +9.1\% [+2.4, +16.9] ($n_T$=27) & +6.8\% [+2.8, +11.3] ($n_T$=103) \\
Headroom v0.27.0 & +47.3\% [+35.2, +59.2] ($n_T$=27) & +48.4\% [+42.3, +55.0] ($n_T$=103) \\
\bottomrule
\end{tabular}
\end{table}

\subsection{Token reduction versus cost reduction}\label{sec:decoupling}

\begin{figure}[t]
\centering
\includegraphics[width=0.52\linewidth]{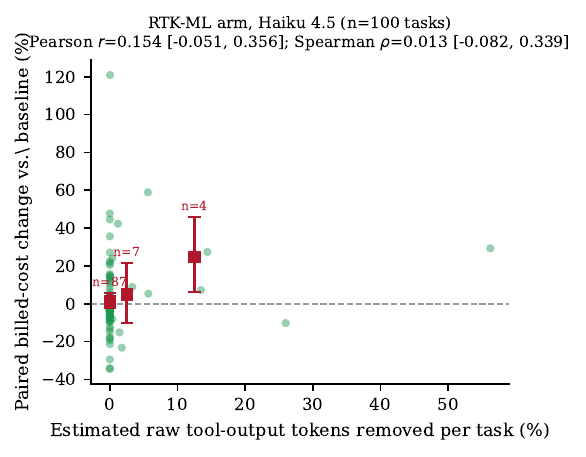}
\caption{Per-task estimated raw tool-output reduction (local BPE
tokenizer) vs.\ paired billed-cost change
(\ninegate{} arm, Haiku 4.5; each point one task, $n=100$; task means over
paired blocks; red squares: binned means with bootstrap 95\% CIs). Pearson
$r=0.154$ \ci{-0.051}{+0.356}; Spearman $\rho=0.013$ \ci{-0.082}{+0.339}
(task bootstrap, 10{,}000 resamples, seed 7). Observed reduction is
endogenous to the arm's own trajectory, not an assigned dose.}
\label{fig:reduction}
\end{figure}

The hook-side ledger makes raw-vs-delivered tool output observable for
the RTK arms. Over the campaign, the \ninegate{} arm reduced estimated raw tool-output tokens from 5.62M
to 3.46M delivered ($-38.4\%$), while RTK delivered essentially raw
output ($-1.3\%$). Raw and delivered counts are produced by RTK's
hook-side ledger using an embedded local BPE tokenizer (\texttt{tiktoken}
\texttt{o200k\_base} merge tables), not the provider tokenizer, so we
report them as estimates; the reduction percentage compares like-for-like
counts under the same tokenizer. Yet the \ninegate{} arm's
paired billed cost was \emph{higher} than baseline and RTK's was
marginally lower. At the task level (\cref{fig:reduction}), observed
tool-output reduction was a weak, unstable predictor of paired billed-cost
change: Pearson $r=0.154$ \ci{-0.051}{+0.356}, Spearman $\rho=0.013$
\ci{-0.082}{+0.339} over 100 Haiku tasks; the Pearson interval crosses
zero, the point estimate rises to $0.24$ when the five highest-reduction
tasks are removed, and it is $0.13$ \ci{-0.09}{+0.36} on the 68 tasks
where the gate actually routed output. Two caveats bound the
interpretation. First, reduction is concentrated: 87 of 100 tasks saw under
0.1\% reduction, and the four tasks in the 5--20\% band show a
\emph{positive} mean cost change ($+24.7\%$ \ci{+6.3}{+46.0}; binned means
in \cref{fig:reduction}). Second, observed reduction is endogenous---the
optimized arm can issue different commands and different trajectories, so
raw-output exposure is partly determined by the model's own actions, and
the compression percentage is not an externally assigned treatment dose. We
therefore read $r$ descriptively: in this evaluated Haiku task set,
per-task observed tool-output reduction was a weak predictor of paired
billed-cost change; this does not establish that compression cannot affect
cost. Answering RQ3, component-level reduction did not predict end-to-end
cost movement in these workloads. The phase analysis (\cref{sec:phases})
identifies trajectory patterns consistent with a possible mechanism: the
realized saving (\cref{eq:realized}) is small once cache prices and added
turns are accounted for---retrieval-stage savings were real but small at
cache prices, and were offset downstream by
added diagnosis, testing, and re-retrieval turns.

\subsection{Cost per successful execution}

\begin{table}[t]
\centering
\caption{Billed-cost change and cost per successful execution (\cref{eq:cps}), with task-bootstrap
95\% CIs. Top: main-campaign estimates for RTK v0.44.1, RTK-ML, and
Headroom, followed by the post-hoc RTK v0.44.1 medium-effort replication
on Opus~4.8 and Opus~5. Bottom: model-level CPS
indices; $n_T$ = unique tasks, s/r = successes/runs
(\textsuperscript{u} = fewer than six tasks; main-campaign Opus~4.8 cells
rest on 38 runs each and are based on smaller samples;
\textsuperscript{r} = post-hoc replication campaign, medium effort,
52 tasks, Claude Code 2.1.220, \cref{app:upstream}). Failed runs remain
in the cost numerator.}
\label{tab:cps}
\small
\setlength{\tabcolsep}{3.2pt}
\begin{tabular}{llccc}
\toprule
System & Scope & $\Delta$ billed cost [95\% CI] & CPS ratio [95\% CI] & Succ./runs \\
\midrule
Baseline Claude Code & main campaign & reference & 1.000 & 685/712 \\
RTK v0.44.1 & main campaign & $-2.7\%$ [-5.6,-0.1] & 0.968 [0.937,1.004] & 686/712 \\
RTK-ML & main campaign & $+6.8\%$ [+2.8,+11.3] & 1.051 [1.006,1.100] & 694/712 \\
Headroom v0.27.0 & main campaign & $+48.4\%$ [+42.3,+55.0] & 1.464 [1.397,1.531] & 696/712 \\
\midrule
RTK v0.44.1$^{r}$ & Opus 4.8 & $\mathbf{-2.9\%}$ [-18.8,+9.3] & 0.933 [0.795,1.073] & 51/52 \\
RTK v0.44.1$^{r}$ & Opus 5 & $\mathbf{+2.5\%}$ [-6.1,+12.0] & 1.025 [0.941,1.123] & 51/52 \\
RTK v0.44.1$^{r}$ & pooled & $-0.1\%$ [-9.3,+7.8] & 0.979 [0.896,1.066] & 102/104 \\
\bottomrule
\end{tabular}
\\[6pt]
\begin{tabular}{llccc}
\toprule
System & Model & CPS index vs.\ model baseline [95\% CI] & $n_T$ & s/r \\
\midrule
Baseline Claude Code & Haiku 4.5 & 1.000 [0.863, 1.156] & 103 & 412/437 \\
Baseline Claude Code & Sonnet 5 & 1.000 [0.886, 1.142] & 103 & 235/237 \\
Baseline Claude Code & Opus 4.8 & 1.000 [0.828, 1.187] & 30 & 38/38 \\
Baseline Claude Code & Opus 5\textsuperscript{r} & 1.000 [0.849, 1.172] & 52 & 51/52 \\
RTK v0.44.1 (main) & Haiku 4.5 & 0.989 [0.858, 1.151] & 103 & 415/437 \\
RTK v0.44.1 (main) & Sonnet 5 & 0.967 [0.873, 1.076] & 103 & 234/237 \\
RTK v0.44.1 (main) & Opus 4.8 & 0.931 [0.797, 1.079] & 30 & 37/38 \\
RTK v0.44.1 (medium) & Opus 4.8\textsuperscript{r} & 0.933 [0.795, 1.073] & 52 & 51/52 \\
RTK v0.44.1 (medium) & Opus 5\textsuperscript{r} & 1.025 [0.941, 1.123] & 52 & 51/52 \\
RTK-ML & Haiku 4.5 & 0.981 [0.855, 1.129] & 103 & 426/437 \\
RTK-ML & Sonnet 5 & 1.096 [0.945, 1.277] & 103 & 231/237 \\
RTK-ML & Opus 4.8 & 1.202 [0.909, 1.553] & 30 & 37/38 \\
Headroom v0.27.0 & Haiku 4.5 & 1.474 [1.277, 1.715] & 103 & 422/437 \\
Headroom v0.27.0 & Sonnet 5 & 1.420 [1.285, 1.570] & 103 & 236/237 \\
Headroom v0.27.0 & Opus 4.8 & 1.673 [1.359, 1.997] & 30 & 38/38 \\
\bottomrule
\end{tabular}

\end{table}

\begin{figure}[t]
\centering
\includegraphics[width=0.7\linewidth]{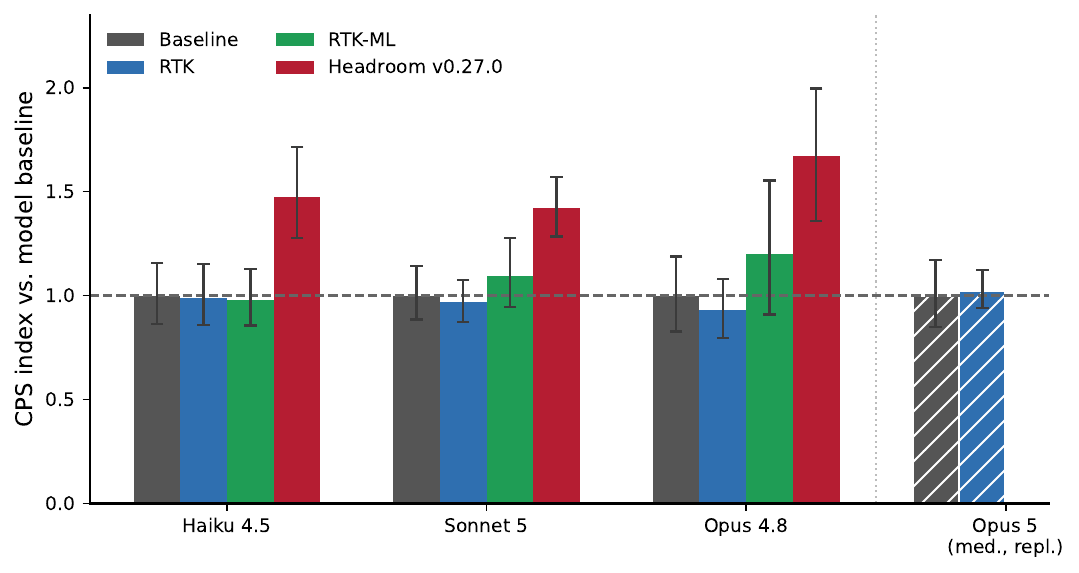}
\caption{Cost per successful execution by system and model (whiskers: task-bootstrap 95\% CIs; 10{,}000 resamples, seed 7; success
and run counts in \cref{tab:cps}). The hatched Opus 5 group is the post-hoc medium-effort RTK v0.44.1
replication cell (CPS 1.025, 95\% CI [0.941, 1.123]); the Opus~4.8
medium-effort replication value is reported in \cref{tab:cps,tab:upstream}.}
\label{fig:cps}
\end{figure}

\begin{figure}[t]
\centering
\includegraphics[width=0.95\linewidth]{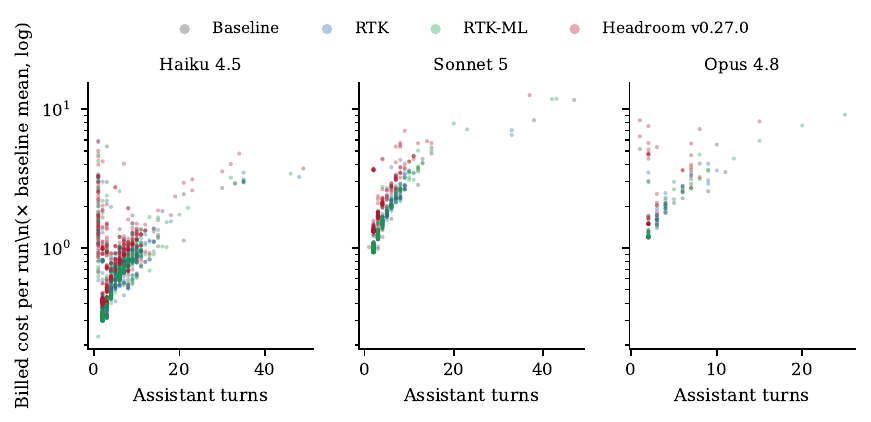}
\caption{Billed cost vs.\ trajectory length (assistant turns) per run, by
model (n=2,848 runs; log-scaled cost). Each added turn re-transmits the
cached prefix, so trajectory changes dominate per-turn savings.}
\label{fig:turnscost}
\end{figure}

\Cref{tab:cps} and \cref{fig:cps} give the primary success-adjusted metric with its
uncertainty. In the main campaign, RTK v0.44.1 has a CPS
ratio of 0.968 \ci{0.937}{1.004}; \ninegate{} is 1.051
\ci{1.006}{1.100} and Headroom is 1.464 \ci{1.397}{1.531}. The post-hoc medium-effort replication uses the same RTK v0.44.1.
There, Opus~4.8 shows a $-2.9\%$ billed-cost
change \ci{-18.8}{+9.3} and CPS ratio 0.933 \ci{0.795}{1.073}; Opus~5
shows a $+2.5\%$ billed-cost change \ci{-6.1}{+12.0} and CPS ratio 1.025
\ci{0.941}{1.123}. Pooled across both post-hoc cells, billed cost changes
by $-0.1\%$ \ci{-9.3}{+7.8} and CPS is 0.979 \ci{0.896}{1.066}. On
every model measured in the main campaign, Headroom's cost per successful
execution is the highest (67\% above baseline on Opus 4.8). \Cref{fig:turnscost} shows the
run-level relationship between trajectory length and billed cost that
drives these aggregates: cost grows with turns on every arm, so any layer
that adds turns must overcome that growth before its per-turn savings show
up as net savings.

\subsection{Single-shot grounded-completion study (separate harness)}\label{sec:grounded}

A companion program ran a \emph{single-shot} SEARCH/REPLACE coding agent
(claude-sonnet-4-6, billed \texttt{anthropic\_api\_usage}) over
SWE-bench-derived Go tasks in the benchmark worktree harness
(\cref{tab:benchprogram}). It is not a Claude Code campaign---one model, one
shot, small $n$---but it is the only place in the artifact record where
\emph{task success under compression} is measured jointly with a grounding
intervention, and it illustrates a broader risk for edit formats that
depend on verbatim anchors.

\paragraph{Compression can alter verbatim edit anchors.} On the full
ContextBench-Go population (40 rows $\times$ \{raw, compressed\}), the
compressed arm cut billable tokens by 73\% and cost by 59\%, yet lost
patch-\emph{applicability} on a net 12 rows (raw applied 27/40, compressed
15/40; failure mix \texttt{edit\_apply\_failed} 20 vs.\ 12): SEARCH/REPLACE
editing requires the agent to reproduce a code anchor byte-for-byte, and
aggressive compression rewrote or dropped exactly those spans. Resolved
tasks were 2/40 (raw) vs.\ 1/40 (compressed), and on the 10-row subset where
both arms applied, solve rate was equal (1--1). The observed deficit was concentrated in
patch applicability; the experiment does not isolate whether compression
also affected reasoning quality. The causal chain for application
failures is mechanical: the
model receives context, emits an edit whose SEARCH block must match the
on-disk file byte-for-byte, and application precedes any test; compression
that rewrites, omits, normalizes, or reorders the anchored span fails the
patch even when the intended change is conceptually correct. The failure
taxonomy therefore separates four states---evidence loss before reasoning,
reasoning failure, edit-application failure, and test failure after
successful application---and the observed failure mix moves between them:
under grounding, \texttt{edit\_apply\_failed} decreased (17$\to$5 on the
compressed arm) and \texttt{tests\_failed} became the dominant residual
(genuine reasoning limits). A context-only agent on
SWE-bench\_Pro Go rows applied 0/18 patches, a pattern consistent with
the same anchor-availability limitation. The frozen
report itself warns against the ``successes-per-million-tokens'' framing
under which the compressed arm would appear superior: with 1 vs.\ 2 solves on
non-overlapping subsets, that ratio metric inverts the verdict while obscuring the reduction in patch-application rate.

\paragraph{Effect of byte-exact grounding.}
Phase G2 re-ran 29 oracle-resolved rows across four arms, adding a
\emph{grounding} prelude---byte-exact $\pm$30-line windows read from the
prepared repository at the gold-context file paths---ahead of the (raw or
compressed) broad context (\cref{tab:groundede2e}). Grounding lifted apply
rates from 76\% to 90\% (raw) and from 31\% to 83\% (compressed) and
produced the program's first nonzero solve-rate improvement: 5/29 for grounded raw at
the lowest observed cost per resolved row among the arms. Under grounding, the compressed arm's solves were a
strict subset of the raw arm's: three tasks solved by the grounded raw arm
were not solved by the grounded compressed arm, while no task showed the
reverse pattern (\cref{tab:g2paired}). With only three discordant pairs the
paired contrast is not statistically resolvable (exact two-sided $p=0.25$),
and per-row costs were not retained, so no confidence interval can be
attached to the per-resolved-row cost ratio; descriptively, the
compressed grounded arm produced fewer solves at a higher observed cost
per resolved row
(2.08$\times$ the raw arm's) even though its cost \emph{per attempted row} was
17\% lower---compression reduced per-attempt spend while
reducing application probability, while observed cost per resolved row increased. These counts are small; we report them as evidence about the mechanism, not as a
stable general result. A
family-routing variant of the same design showed the preserve policy
(\cref{sec:architectures}, gate 7) preserving this content type:
preserve vs.\ aggressive compression applied 83\% vs.\ 48\% context-only,
with evidence-loss failures 5 vs.\ 15, and grounded-preserve solving most
(6) at the lowest observed cost per resolved row of the family study.

\begin{table}[t]
\centering
\caption{Phase G2 grounded-completion results (29 oracle-resolved
ContextBench-Go rows per arm, single-shot, claude-sonnet-4-6, billed usage;
token-reconstructed costs reported as indices). In this small study, the largest observed improvement was
associated with grounding; under grounding, the compressed arm produced
descriptively fewer solves at higher observed cost per resolved row
(\cref{tab:g2paired}).}
\label{tab:groundede2e}
\small
\begin{tabular}{lcccc}
\toprule
Arm & Applied & Resolved & Billable tok & Cost index \\
\midrule
raw context-only & 22/29 (76\%) & 1 (3.4\%) & 114,658 & 1.00$\times$ \\
compressed context-only & 9/29 (31\%) & 1 (3.4\%) & 37,157 & 0.45$\times$ \\
grounded raw & 26/29 (90\%) & 5 (17.2\%) & 347,231 & 2.18$\times$ \\
grounded compressed & 24/29 (83\%) & 2 (6.9\%) & 276,531 & 1.82$\times$ \\
\bottomrule
\end{tabular}
\end{table}

\begin{table}[t]
\centering
\caption{Paired task outcomes and cost decomposition in the grounded study
(29 rows; grounded raw vs.\ grounded compressed). With three discordant
pairs, the exact two-sided $p$ is 0.25; the comparison is descriptive.
Per-row costs were not retained, so uncertainty for the per-resolved-row
cost cannot be computed.}
\label{tab:g2paired}
\small
\begin{tabular}{lc}
\toprule
Paired outcome (29 rows) & Count \\
\midrule
Solved by both arms & 2 \\
Solved only by grounded raw & 3 \\
Solved only by grounded compressed & 0 \\
Solved by neither & 24 \\
\midrule
Exact two-sided $p$ (discordant pairs) & 0.25 \\
\midrule
Cost per attempted row (compressed vs.\ raw) & 0.83$\times$ raw \\
Cost per applied patch (compressed vs.\ raw) & 0.90$\times$ raw \\
Cost per resolved row (compressed vs.\ raw) & 2.08$\times$ raw \\
\bottomrule
\end{tabular}
\end{table}

\paragraph{Scope limits.} These are single-shot trials: the agent cannot
search again or recover, so the recovery-loop dynamics of
\cref{sec:phases} are structurally invisible here (the IC-SWE pilot records
them explicitly as \texttt{not\_measured}). Sample sizes are small (solve
counts of five and two), one model (Sonnet 4.6) and one edit format
(SEARCH/REPLACE) were used, tasks are SWE-bench-derived Go rows, and one
compression configuration was evaluated. These results should not be generalized beyond this experimental setting. We interpret this program as evidence about the mechanism---what the
evaluated compression configuration does to edit anchors, and what
grounding does to apply rates---not as a system ranking; its direction is
consistent with the paired multi-turn campaigns.

\subsection{Long-session evidence (separate lineage)}\label{sec:largesessions}

A separate long-session program (Opus 4.8; three long-horizon
code-change tasks on synthetic large repositories; paired design) provides
the long-session view: a 529-session main campaign, preceded by a
superseded 133-session two-system generation and followed by a 60-session
attribution follow-up, all separately
billed (disjoint session identifiers), with the caveat that its RTK ``current'' arm was built from an earlier
binary lineage (the pre-RTK-ML observer build), so its results must not
be merged with \cref{tab:aggregate}. There, RTK was cost-neutral
($-1.3\%$, n.s.), the observer-lineage build was $6.5\%$ \emph{more}
expensive (CI \ci{1.5}{11.7}), and Headroom was $60.9\%$ more expensive,
with 131/132 task success. The qualitative pattern---no consistent cost reduction from compression on long sessions and substantial proxy overhead---is consistent with the authoritative campaign.

\section{Heterogeneity by Model, Effort, and Task Family}\label{sec:heterogeneity}

\subsection{Model $\times$ effort}

\begin{table}[t]
\centering
\begin{threeparttable}
\caption{Model $\times$ effort matrix: paired billed-cost change vs.\
baseline (\% of same-cell baseline mean; task-clustered bootstrap 95\% CI;
$n$ = paired tasks). Cells never measured in any campaign are stated as
such, not imputed.}
\label{tab:modeleffort}
\scriptsize
\setlength{\tabcolsep}{2.6pt}
\begin{tabular}{llccc}
\toprule
Model & Effort & RTK v0.44.1 $\Delta$\% & RTK-ML $\Delta$\% & Headroom $\Delta$\% \\
\midrule
Haiku 4.5 & low & +7.0 [-2.7, +19.1] (76) & +6.8 [-2.0, +18.3] (76) & +55.3 [+40.6, +72.8] (76) \\
Haiku 4.5 & medium & -3.8 [-14.9, +5.9] (20)\textsuperscript{u} & +32.9 [+6.2, +69.5] (20)\textsuperscript{u} & +110.2 [+41.1, +197.8] (20)\textsuperscript{u} \\
Haiku 4.5 & high & -2.6 [-11.5, +5.7] (103) & -1.6 [-12.5, +7.3] (103) & +47.9 [+34.6, +62.3] (103) \\
Haiku 4.5 & xhigh & -10.5 [-31.4, +2.5] (20)\textsuperscript{u} & -8.7 [-50.6, +20.1] (20)\textsuperscript{u} & +46.4 [+28.2, +66.2] (20)\textsuperscript{u} \\
Haiku 4.5 & max & -9.9 [-32.2, +11.8] (20)\textsuperscript{u} & -21.2 [-40.6, -4.8] (20)\textsuperscript{u} & +9.0 [-13.6, +30.6] (20)\textsuperscript{u} \\
Sonnet 5 & low & +10.3 [-3.4, +32.8] (20)\textsuperscript{u} & +1.2 [-5.7, +8.1] (20)\textsuperscript{u} & +80.8 [+49.0, +116.9] (20)\textsuperscript{u} \\
Sonnet 5 & high & -4.9 [-10.9, +0.1] (103) & +8.3 [+3.0, +14.4] (103) & +39.5 [+32.9, +46.5] (103) \\
Opus 4.8 & low & -2.8 [-13.7, +7.3] (8)\textsuperscript{u} & +34.8 [-6.2, +109.6] (8)\textsuperscript{u} & +81.1 [+41.3, +123.8] (8)\textsuperscript{u} \\
Opus 4.8 & medium\textsuperscript{r} & -2.9 [-18.8, +9.3] (52) & \multicolumn{2}{c}{\emph{not arms of the replication}} \\
Opus 4.8 & high & -11.0 [-23.1, -2.2] (30) & +12.4 [-6.8, +34.7] (30) & +63.7 [+37.5, +91.7] (30) \\
Opus 5 & medium\textsuperscript{r} & +2.5 [-6.1, +12.0] (52) & \multicolumn{2}{c}{\emph{not arms of the replication}} \\
Sonnet 5 & med./xhigh/max & \multicolumn{3}{c}{\emph{not measured in any campaign}} \\
Opus 4.8 & xhigh/max & \multicolumn{3}{c}{\emph{not measured in any campaign}} \\
Opus 5 & low/high/xhigh/max & \multicolumn{3}{c}{\emph{not measured in any campaign}} \\
\bottomrule
\end{tabular}
\begin{tablenotes}\footnotesize
\item \textsuperscript{u} exploratory stratum ($n{=}20$ tasks or fewer;
the campaign's power analysis flags such cells and the expansion analysis
found arm$\times$effort interactions not significant).
\item \textsuperscript{r} post-hoc RTK v0.44.1 replication
(\cref{app:upstream}): 52 family-stratified tasks, medium effort,
Claude Code 2.1.220. RTK-ML and Headroom were not arms of that campaign.
\end{tablenotes}
\end{threeparttable}
\end{table}

\begin{figure}[t]
\centering
\includegraphics[width=0.86\linewidth]{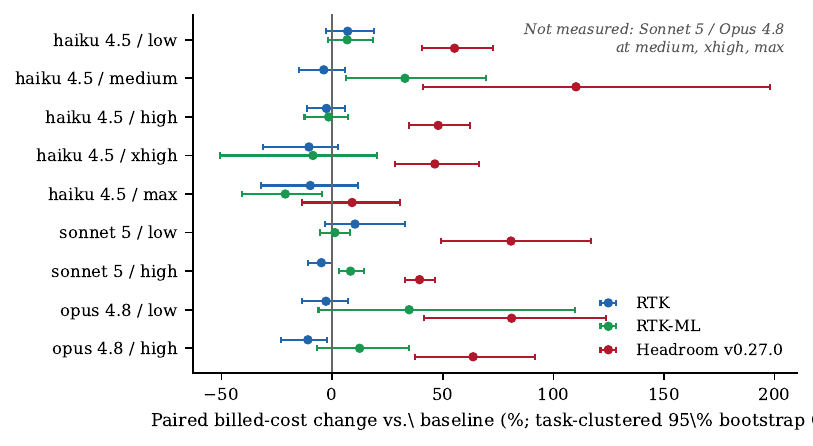}
\caption{Paired billed-cost change vs.\ baseline by main-campaign model--effort cell
(task-clustered 95\% CIs). This figure shows the main campaign only; the
post-hoc RTK v0.44.1 Opus-medium cells are reported in \cref{tab:modeleffort,tab:upstream}.
Unmeasured main-campaign cells are annotated, not imputed.}
\label{fig:modeleffort}
\end{figure}

\Cref{tab:modeleffort} and \cref{fig:modeleffort} present all nine
main-campaign model--effort cells, plus two post-hoc medium-effort RTK
v0.44.1 cells (Opus 4.8 and Opus 5) contributed by the replication
campaign. Three observations are relevant. First, Headroom's cost increase
is positive in every observed cell ($+9.0\%$ to $+110.2\%$). Second,
RTK effects hover in single digits and flip sign between cells; the
expansion campaign's interaction tests found neither arm$\times$model nor
arm$\times$effort effects significant, and specifically retired the
apparent effort flips---e.g., the \ninegate{} Haiku-medium cell at
$+32.9\%$ \ci{+6.2}{+69.5} against the Haiku-max cell at $-21.2\%$---as
small-sample artifacts of 20-task exploratory strata. We therefore do not
promote any effort-conditioned deployment rule. Third, coverage is
incomplete by design and budget: Sonnet 5 was never run at medium, xhigh,
or max effort in any campaign, and Opus 4.8 at xhigh or max; the medium-effort Opus cells shown in the model--effort table are the
post-hoc RTK v0.44.1 replication results. All other cells are absent from
every figure and table rather than estimated.

\subsection{Task families}

\begin{table}[t]
\centering
\caption{Task-family effects for families with $\geq$3 analyzed tasks:
paired billed-cost change vs.\ baseline (\% of family baseline mean;
task-clustered bootstrap 95\% CI). One- and two-task families are
exploratory and reported in Appendix~\ref{app:familyfull}. Families are
descriptive decompositions, not pre-specified contrasts.}
\label{tab:family}
\small
\setlength{\tabcolsep}{4pt}
\begin{tabular}{lrccc}
\toprule
Task family & $n_T$ & RTK $\Delta$\% & RTK-ML $\Delta$\% & Headroom $\Delta$\% \\
\midrule
lexical\_search & 19 & -0.7 [-4.1, +2.4] & -2.3 [-5.2, -0.3] & +63.0 [+48.7, +79.4] \\
symbol\_location & 14 & +1.2 [-1.7, +4.6] & +3.6 [-0.7, +8.6] & +51.9 [+37.3, +69.9] \\
dependency\_tracing & 13 & -6.7 [-15.2, -0.2] & +1.9 [-4.1, +8.2] & +60.2 [+41.3, +80.2] \\
semantic\_search & 11 & -3.2 [-12.6, +5.2] & +12.4 [-1.4, +28.2] & +50.7 [+28.9, +69.1] \\
architecture & 7 & +1.3 [-7.2, +9.3] & -0.8 [-6.6, +6.0] & +57.1 [+33.1, +86.5] \\
long\_session & 6 & -4.0 [-9.6, +3.0] & +4.7 [-9.5, +16.0] & +44.6 [+24.2, +65.1] \\
multi\_file\_debug & 4 & -9.7 [-20.7, +1.4] & +25.0 [-2.2, +52.2] & +37.7 [+12.5, +64.5] \\
feature\_impl & 3 & +3.9 [-9.9, +17.4] & +7.3 [-9.6, +21.6] & +36.2 [+31.6, +40.0] \\
negative\_control & 3 & -1.5 [-2.4, -0.6] & -2.9 [-6.9, -0.7] & +81.7 [+30.7, +147.1] \\
root\_cause\_log & 3 & -2.1 [-3.3, -0.1] & +2.3 [-2.1, +9.3] & +46.4 [+21.4, +95.1] \\
shell\_pipeline & 3 & -1.5 [-2.1, -0.7] & -1.0 [-2.1, +0.2] & +50.9 [+36.5, +79.6] \\
structured\_data & 3 & -1.0 [-10.6, +4.5] & +7.7 [-4.9, +31.7] & +43.0 [+31.9, +62.4] \\
\bottomrule
\end{tabular}
\end{table}

Across the 23 analyzed families (12 with $\geq$3 tasks in
\cref{tab:family}; the remainder in Appendix~\ref{app:familyfull};
\cref{fig:family} shows all), Headroom's cost delta is positive in
\emph{all} of them (range $+14.7\%$ to $+82.4\%$). RTK effects are family-heterogeneous: RTK shows its
strongest savings on dependency tracing and multi-file debugging;
\ninegate{} saves modestly on bulk lexical retrieval and shell pipelines
while increasing cost on semantic search and multi-file debugging. The
double-digit \ninegate{} cells are all one-to-four-task families and are
labeled exploratory. Session length is the strongest task-side moderator of
cost composition (\cref{tab:coststack}); the expansion analysis found
retrieval-mode and output-band interactions not significant. The overall
answer to RQ4 is that system effects are \emph{composition-dependent and
heterogeneous}: the earlier campaign's lower cost for \ninegate{} on the weaker-model, bulk-retrieval-heavy mix did not generalize as a model-monotonic pattern on the broader read-only mix.

\begin{figure}[t]
\centering
\includegraphics[width=0.9\linewidth]{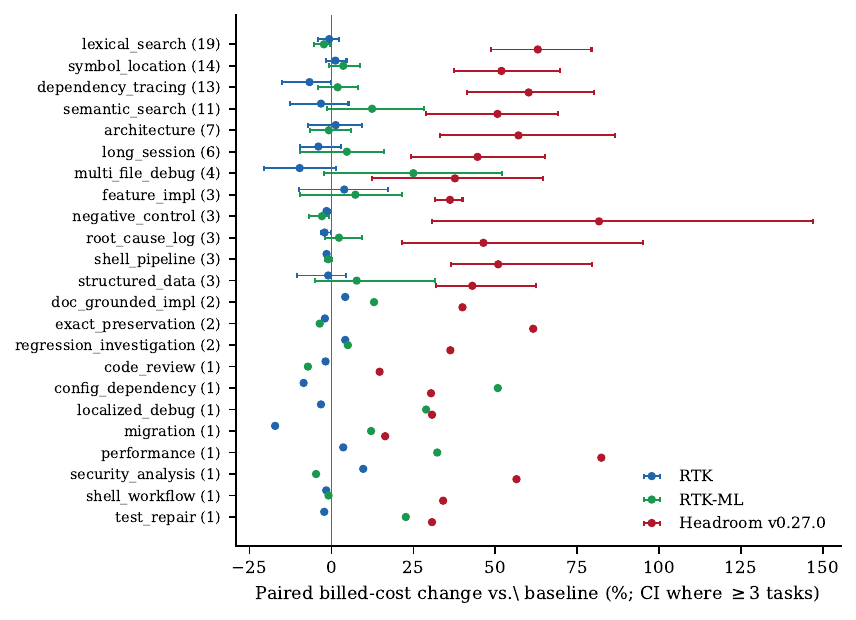}
\caption{Paired billed-cost change by task family (task counts in
parentheses; CIs where $\geq$3 tasks). Headroom is right of zero in every
family; RTK effects change sign across families.}
\label{fig:family}
\end{figure}

\section{Cross-Agent Replication on Codex}\label{sec:codex}

A subsequent replication campaign
(\texttt{codex\_headroom\_replication40\_20260715\_v1}) tested whether the
Headroom cost signal transfers to a different agent harness: the Codex CLI
with \texttt{gpt-5.6-luna} at low reasoning effort. All numbers in this
section are transcribed from that campaign's frozen replication report; its
artifacts (frozen task manifest, sources lock, budget ledger, per-pair
results, raw run telemetry) are retained in the Codex campaign tree, which
is separate from this paper's artifact release. Provider-billed cost is
unavailable in Codex telemetry, so costs here are \emph{reconstructed from
provider-returned token counts at list prices}---evidence class
``tokens,'' not ``billed,'' in the terms of \cref{tab:benchmatrix}.

\paragraph{Design.} Forty tasks were sampled deterministically (seed
20260715) from the frozen 110-task pool after excluding all 12 tasks of
the original Codex pilot, giving a disjoint sample. Two arms---baseline
and a fixture-preserving, isolated Headroom-only configuration---ran in
randomized order within every paired block: 80 new paid runs in 40
complete pairs, with zero paid retries. Benchmark commit, tasks, prompts,
graders, fixtures, and success criteria were unchanged from the pilot.
Activation evidence is complete: all 40 Headroom runs retained a live
proxy WebSocket request, and all 40 baseline runs were isolated from the
proxy.

\begin{table}[H]
\centering
\caption{Codex replication, Headroom-only vs.\ baseline (40 disjoint task
pairs; costs reconstructed from provider token counts; differences are
Headroom relative to baseline). Transcribed from the frozen replication
report.}
\label{tab:codex}
\small
\begin{tabular}{lccc}
\toprule
Metric & Baseline & Headroom-only & Difference \\
\midrule
Task success & 39/40 & 39/40 & 0 discordant pairs \\
Reconstructed cost & --- & --- & $-12.49\%$ \\
Cost per successful task & --- & --- & $-12.49\%$ \\
Input tokens & --- & --- & $-10.88\%$ \\
Cached input tokens & --- & --- & $-8.33\%$ \\
Output tokens & --- & --- & $+6.63\%$ \\
Reasoning tokens & --- & --- & $+31.30\%$ \\
Completed shell calls & 71 & 76 & $+7.04\%$ \\
Delivered shell-output bytes & --- & --- & $+3.67\%$ \\
Wall time & --- & --- & $+238.52\%$ \\
\bottomrule
\end{tabular}
\end{table}

\paragraph{Result.} Both arms passed 39/40 tasks with no discordant
correctness outcome (both failed the same task, returning the same
incorrect frozen-fixture subject). Headroom's reconstructed cost was
12.49\% lower than baseline, and identically so per successful task. The
paired per-task change averaged $-14.76\%$ (median $-19.49\%$); Headroom
was cheaper on 35 of 40 tasks. The saving came from lower input and
cached-input token volumes---\emph{not} from measured output
compression: Headroom again reported zero compressed tokens across all
runs, so the difference is an observed proxy/cache/trajectory effect, not
a demonstrated consequence of its compression counter. The lower reconstructed cost was accompanied by higher latency: wall time more than tripled ($+238.5\%$), and
output tokens, reasoning tokens, shell calls, and delivered shell-output
bytes all increased.

\paragraph{Combined Codex evidence.} Pooling this disjoint replication
descriptively with the original 12-task pilot gives 52 task pairs: 51/52
success on both arms and a $-14.35\%$ reconstructed-cost difference. The
larger sample strengthens the association between this Headroom
configuration and lower reconstructed Codex cost at equal task success; it
does not establish that compression caused the saving, and latency remains
a severe regression.

\paragraph{Cross-campaign 12-task slice: Headroom vs.\ RTK.} The original
12 pilot tasks can be matched by task identifier to the same tasks in a
later patched-RTK Codex campaign. This is a cross-campaign comparison---
each treatment has its own concurrent baseline, not a randomized
simultaneous three-arm design. Against its own baseline, Headroom was
cheaper by 19.53\% (cheaper on 11/12 tasks, 12/12 success); RTK was
cheaper by 10.45\% (cheaper on 5/12, 11/12 success---its one failure was
a known counting regression in which compacted output changed a required
count). Within this matched slice Headroom shows the larger observed cost reduction, but the
cross-campaign design prevents attributing the difference solely to
treatment. The RTK slice also has a distributional limitation: the mean reduction reflects a minority of tasks (cheaper on 5 of 12, more expensive on the rest), so the mean does not represent a stable per-task reduction. Distribution-level analysis on Codex---median and
tail behavior, and whether the improving minority is identifiable in
advance---remains open.

\paragraph{Integrity notes.} Eight pairs initially returned a grader
exit-127 because the campaign judge PATH exposed only \texttt{python3};
no paid run was repeated---all 16 retained worktrees passed the
unchanged graders offline under the pinned test environment. One baseline
telemetry heuristic flagged the word ``headroom'' in the campaign's
directory path; preflight configuration, commands, environment, and zero
proxy requests establish this as a path-name false positive. One local
Headroom process failed before proxy readiness and before any API request;
the failure record is retained with the campaign artifacts, and the
installation was repaired from the same pinned commit before the paid
campaign began.

\paragraph{Interpretation.} On Claude Code, the tested Headroom
configuration was the study's largest cost increase (CPS ratio 1.464); on
Codex, an isolated Headroom-only configuration was associated with a
12--14\% reconstructed-cost \emph{saving} at equal success. The sign
reversal across harnesses, with zero compressor activity in both, is
additional evidence for the paper's main conclusion: context-layer effects are
properties of the layer--harness--model--workload combination, not of the
layer alone, and must be measured end to end in each deployment.

\section{Trajectory and Phase Efficiency}\label{sec:phases}

\subsection{Phase taxonomy and validation}

Every retained transcript of the base campaign was decomposed into
assistant turns and each turn assigned one of eight phases (orientation,
retrieval, diagnosis, planning, implementation, testing/debug,
verification, final response) plus an ambiguous/mixed catch-all, using a
frozen deterministic priority rule (edits win over tests, tools over text;
no model calls). The ledger covers 13,620 classified turns from 1,120 base
campaign runs. A manual implementation audit independently reproduced 40/40 sampled rule
assignments, and 2.0\% of turns (187/9,523 in the audited slice) fell into
the ambiguous class. The audit validates that the deterministic rules were
applied as specified; it does not establish that the labels match
independent human judgments of cognitive phase. Mixed tool-using turns are
forced into mutually exclusive categories, so the taxonomy characterizes
observable turn behavior, not latent reasoning; a blind semantic validation
by independent annotators was not performed and is listed as a limitation. The near-zero planning share is primarily a property
of the frozen classifier, which assigns mixed tool-using turns to other
phases; it should not be read as evidence that planning is absent from
coding agents. Expansion-split phase claims require regeneration from the
retained transcripts, which has not been performed; all phase numbers here
are base-campaign numbers.

\subsection{Where generated tokens go}

\begin{table}[t]
\centering
\caption{Share of generated (output) tokens by phase and arm
(characterization split; 10{,}376 turns). Retrieval plus diagnosis account
for $\approx$58--61\% of generated tokens on every arm.}
\label{tab:phaseshares}
\small
\begin{tabular}{lcccc}
\toprule
Phase & Baseline & RTK & RTK-ML & Headroom \\
\midrule
P1 orientation & 6.3 & 6.2 & 5.6 & 5.0 \\
P2 retrieval & 35.1 & 35.4 & 34.5 & 36.5 \\
P3 diagnosis & 23.7 & 23.8 & 24.1 & 24.5 \\
P4 planning & 0.0 & 0.1 & 0.0 & 0.0 \\
P5 implementation & 27.0 & 26.6 & 28.8 & 25.6 \\
P6 testing/debug & 1.3 & 1.3 & 1.7 & 1.2 \\
P7 verification & 0.8 & 0.9 & 1.1 & 0.8 \\
P8 final response & 3.3 & 3.3 & 2.7 & 3.2 \\
ambiguous/mixed & 2.5 & 2.5 & 1.4 & 3.3 \\
\bottomrule
\end{tabular}
\end{table}

\begin{figure}[t]
\centering
\includegraphics[width=0.88\linewidth]{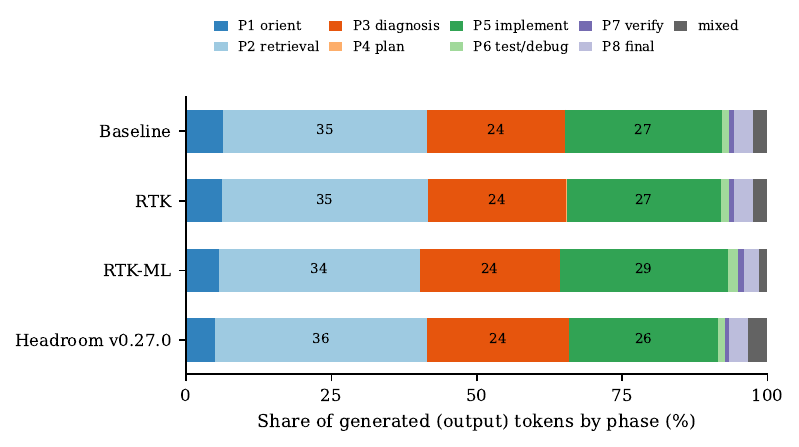}
\caption{Generated-token composition by phase (characterization split;
10{,}376 turns). Coarse phase-token shares were similar across arms
under the frozen turn-level taxonomy.}
\label{fig:phase}
\end{figure}

\Cref{tab:phaseshares} and \cref{fig:phase} show that coarse phase-token shares were similar across arms under the frozen
turn-level taxonomy: retrieval $\approx$35\%
(33--36\%), diagnosis $\approx$24\%, implementation 26--29\%. Retrieval
plus diagnosis account for approximately 58\% of generated tokens
(58.8\%/59.2\%/58.6\%/60.9\% across the four arms). Within this campaign and under the frozen turn-level taxonomy, the arms
differed less in coarse phase composition than in the measured cost and
context volume associated with those phases (delivered tool bytes and the
cache traffic they induce).

\subsection{Phase-level cost changes}

\begin{table}[t]
\centering
\caption{Phase-resolved paired cost deltas vs.\ baseline (percent of the baseline mean run cost;
billed cost allocated to turns by price-weighted token usage; base-campaign
phase ledger). The RTK arms save in retrieval (P2); the \ninegate{} arm
shows downstream cost increases that offset the retrieval-phase reduction.}
\label{tab:waterfall}
\small
\begin{tabular}{lccc}
\toprule
Phase & RTK $\Delta$ & RTK-ML $\Delta$ & Headroom $\Delta$ \\
\midrule
P1 orientation & +0.53 & +1.39 & +11.40 \\
P2 retrieval & -3.07 & -1.96 & +18.14 \\
P3 diagnosis & -0.99 & +4.83 & +17.81 \\
P4 planning & +6.19 & -0.29 & -3.29 \\
P5 implementation & -0.15 & +7.87 & +0.51 \\
P6 testing/debug & +0.24 & +10.94 & +12.72 \\
P7 verification & +0.22 & +3.57 & +3.51 \\
P8 final response & -0.20 & -0.29 & +3.59 \\
ambiguous/mixed & +1.68 & +6.77 & +15.15 \\
\bottomrule
\end{tabular}
\end{table}

The phase-resolved cost decomposition (\cref{tab:waterfall}) shows both
the RTK arms first diverging from baseline at the retrieval phase with lower cost, while Headroom diverges at orientation already---its
higher cache-read usage is concentrated in the early phases (62\% of it accumulates before the
first edit) and reaches roughly $+180$k cache-read
tokens per run by the final phase. For the \ninegate{} arm, downstream cost increases are approximately four times the retrieval-phase reduction: added diagnosis,
implementation, testing/debug, and ambiguous-turn cost, a later first edit
($+0.94$ turns), more post-edit retrieval re-entries ($+0.32$/run), and
more repeated-search tokens ($+301$/run). RTK is the only arm with
downstream behavior closest to baseline and slightly \emph{earlier} first edits
($-0.60$ turns, $-0.26$ re-entries). These are observed associations from
the frozen ledgers, not causal decompositions.

\begin{figure}[t]
\centering
\includegraphics[width=0.82\linewidth]{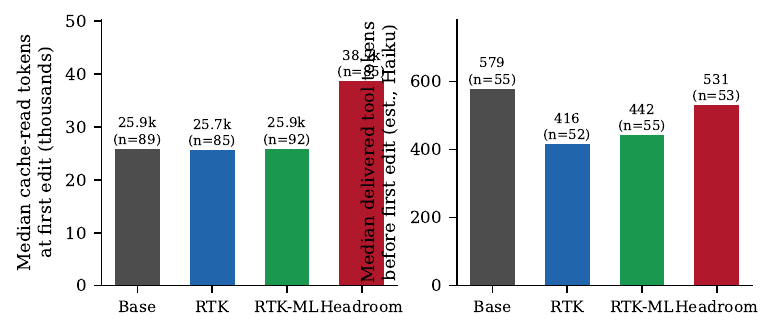}
\caption{Left: median cache-read context at the first code edit (runs with
an implementation phase, base splits, all models). Right: median estimated
delivered tool tokens before the first edit (Haiku runs; bytes/4 estimate).
Headroom has $\approx$49\% more cache-read context at the first edit; the RTK arms
match baseline context at the same milestone.}
\label{fig:preedit}
\end{figure}

Two milestone measurements provide additional detail (\cref{fig:preedit}). Median
cache-read context at the first code edit is essentially identical for
baseline and both RTK arms ($\approx$25.7--25.9k tokens) but 49\%
higher for Headroom (38.7k). And on Haiku---the model where the
\ninegate{} compression surface is largest---the median estimated delivered tool-token count before the first edit
was 442 for the \ninegate{} arm versus 579 for baseline, approximately
24\% lower, with heavy upper tails (a frozen campaign report
characterized the reduction as roughly twofold, but its exact metric,
population, and denominator could not be reconstructed from the retained
artifacts, so we report only the auditable per-run ledger median), while
on Sonnet the same
arm reached milestones $\approx$0.4 turns later. On Haiku, reduced delivered retrieval was achievable without a later first-edit milestone, but it did not translate into net billed savings after trajectory effects were included.

\section{Failure Modes and Safety Constraints}\label{sec:failures}

\paragraph{Pipeline incompatibility with downstream shell consumers.} One safety-relevant failure mode in the artifact record was discovered in the first paid
gate campaign: when retrieval routing was enabled, \emph{every} genuine
gates-on task failure (4 of 4) followed the same pattern---the agent piped
proxied search output into a shell counting pipeline
(\texttt{grep~\ldots~|~cut~-d:~-f1~|~sort~|~uniq~-c}), which then parsed
the ranked \emph{search-group summary} instead of raw
\texttt{file:line:content} matches and produced invalid answers (e.g.,
\texttt{src/8} instead of a file name). Tool tokens on that campaign fell
40.3\% while gates-on success fell to 93.3\% against 100\% for every other
arm: an example in which lower tool-token counts coincided with lower task success because the transformed output was incompatible with the downstream consumer. The fix marks a rewritten command segment as a
\emph{stdout consumer} whenever its stdout feeds a pipe or a stdout
redirect; the proxy then skips ranking for that segment (output stays
byte-preserving) and records the skip in the gate trace. Stderr and stdin
redirects still rank; command substitution was already never rewritten. In
subsequent campaigns the guard suppressed 42\% (Haiku) and 30\% (Sonnet)
of otherwise-eligible routes, and in the authoritative campaign it recorded
339/150/28 skips against 521/208/26 routes on Haiku/Sonnet/Opus. The corrected reruns superseded the earlier aggregate result. The observed failure motivates a content-dependent policy: dense, task-critical evidence such as tracebacks, test output, structured shell streams, and patch anchors should be handled differently from redundant retrieval context.

\paragraph{Edit-anchor alteration under compression.} The grounded-completion study
(\cref{sec:grounded}) documents a second failure mode: aggressive
compression of retrieved code strips the byte-exact spans that
SEARCH/REPLACE patch application depends on, reducing patch application
by approximately 44\% (27/40 $\to$ 15/40; 83\% vs.\ 48\% in the family
study) despite a large reduction in token metrics ($-73\%$ billable). The mitigation that worked was
content-aware: the task-family preserve gate bypasses compression for
high-recall and grounded-editing families, and byte-exact grounding windows
restore anchors outright (apply 31\% $\to$ 83\% on the compressed arm).

\paragraph{Other observed failure modes.} (i)~\emph{Over-compression followed by additional agent work}: removed context that the model still needed returned
as repeated searches and diagnosis re-reads (\cref{sec:phases}); the additional searches and diagnosis turns incur normal trajectory costs. (ii)~\emph{Early failure can have low measured cost}: a run that fails quickly can cost less than successful runs; all aggregates here therefore report success alongside cost and use
cost per successful execution for decisions. (iii)~\emph{Inactive code paths can invalidate comparisons}: a free pre-campaign check proved the gate environment
variables are inert in RTK (byte-identical output on the covered
test paths, no gate trace) and
that the gate build changes bytes only when enabled (11,050\,B raw
\texttt{grep} vs.\ 2,570\,B search-group on the fixture)---without such
activation proofs, an ``optimized'' arm can be equivalent to the baseline without detection.
(iv)~\emph{Component benchmarks can encourage unsafe optimization}: the
pytest-rule example in \cref{sec:componentgap} shows a preservation gate
catching a compression rule that scored well on savings while dropping
task-critical markers.

\paragraph{Gate activation summary.} \Cref{tab:gates} condenses the
per-gate activation evidence: every gate is off by default, and the
disabled paths covered by the dedicated equivalence tests were
byte-equivalent; the embedding, structural, and focus-cache paths were
enabled and operational in the authoritative campaign (zero sidecar outages),
but no task was identified in which any of them was uniquely
decisive---activation is not effect.

\subsection{Performance degradation under context reduction}
\label{sec:degradation}

The observed performance-degradation evidence, ordered by directness, is:
\begin{enumerate}
\item \textbf{Binary task-success degradation.} The pre-guard Stage-1
  campaign: 93.3\% vs.\ 100\% success (4/4 failures from pipeline
  incompatibility). The grounded study: three tasks solved by grounded raw were
  not solved by grounded compressed, none reverse (exact paired $p=0.25$;
  descriptive). The authoritative campaign detected no aggregate
  success-rate degradation, but baseline success was 96--98\%, limiting sensitivity to small differences, and the study was not a
  non-inferiority design; its intervals still permit small losses or gains.
\item \textbf{Patch-application degradation.} 27/40 $\to$ 15/40 applies
  under compression (Phase E); 83\% vs.\ 48\% preserve-vs-aggressive in
  the family study; \texttt{edit\_apply\_failed} 20 vs.\ 12.
\item \textbf{Evidence/marker degradation.} The pre-fix pytest rule
  scored 31.9\% savings at only 56.1\% marker recall; ContextBench
  one-size-aggressive reached 39.8\% savings with substantial context
  loss on 45 of 50 conversations.
\item \textbf{Trajectory degradation.} \ninegate{}: later first edit
  ($+0.94$ turns), more post-edit retrieval re-entries ($+0.32$/run), more
  repeated-search tokens ($+301$/run), downstream phase-cost increase
  (\cref{tab:waterfall}).
\item \textbf{Structured-consumer incompatibility.} The stdout-pipeline
  failure above: compression altered program-consumed shell-output semantics,
  not model comprehension.
\item \textbf{Unmeasured quality dimensions.} Code maintainability, patch
  quality beyond test outcomes, partial correctness, latent bug
  introduction, reasoning quality, and human review effort were not
  measured in any campaign; success here is deterministic test/judge
  outcome only.
\end{enumerate}

\begin{table}[t]
\centering
\caption{Gate-activation matrix for RTK-ML's nine gates (condensed). ``Fired'' cites the
authoritative campaign traces (routes/query-propagations/guard-skips on
Haiku/Sonnet/Opus: 521/208/26, 916/395/57, 339/150/28; all comparator arms
0/0/0).}
\label{tab:gates}
\small
\setlength{\tabcolsep}{4pt}
\resizebox{\textwidth}{!}{%
\begin{tabular}{llllp{4.6cm}}
\toprule
Gate & Default & Fired in E2E & Output-changing & Observed effect \\
\midrule
1 Query propagation & off & yes (916/395/57) & no (signal only) & enables gates 2--5 \\
2 Retrieval routing & off & yes (521/208/26) & yes (search-group) & primary compression surface \\
3 Query-aware ranking & off & yes & yes (reorder) & with gates 4--5 \\
4 Lexical leg & off & yes & yes (reorder) & deterministic scoring \\
5 Embedding leg (BGE) & off & active, healthy & in principle & no decisive per-task signature \\
6 Task-family router & off & active & no (routing var) & guards BGE exposure \\
7 Family preserve & off & active & yes (raw bypass) & protects exact-byte tasks \\
8 Structural (ast-grep) & off & active & indirect (hints) & no measurable signature \\
9 Focus cache & off & active & no (latency only) & no measurable signature \\
--- Safety router & \textbf{enforce} & on & no (this layout) & future-guards sensitive output \\
--- Stdout-consumer guard & auto w/ gate 2 & yes (339/150/28 skips) & prevents downstream incompatibility & fixes \cref{sec:failures} failure \\
\bottomrule
\end{tabular}
}
\end{table}

\section{Limits of Component-Level Benchmarks}\label{sec:componentgap}

The artifact record contains component-level results for every retrieval
and compression approach studied; none of them, alone, would have predicted
the end-to-end outcomes of \cref{sec:results}.

\paragraph{ContextBench (gold-context survival).} On the 50-conversation
policy study, an oracle per-task-type policy achieved 12.3\% savings at
95.2\% recall with 0/50 severe drops; a one-size conservative policy
achieved 7.1\% at 95.7\% recall with 7/50 severe failures; and a
one-size aggressive policy reached 39.8\% savings with
substantial context loss on 45 of 50 conversations. The relevant comparison is oracle-vs-fixed rather than the absolute numbers: safe savings are
policy-dependent and modest. These are retrieval-quality proxies; the study
records no task success.

\paragraph{LoCoEval (long-history preservation).} The retained runs are a
preservation-only proxy (reference-answer token survival after compressing
long conversations); the pipeline's LLM-judge stage was unavailable, and
token savings above the counting API's size cap fall back to length/4
estimates. We cite it only as evidence that deterministic command-level
compression does not affect conversational history, which it was not designed to modify,and that long-history compression requires its own track.

\paragraph{InterCode-proxy (dense diagnostic evidence).} On 24 pytest
observations (12 MBPP tasks $\times$ pass/bug), the original deterministic
pytest rule scored 31.9\% savings at only 56.1\% marker recall---dropping
task-critical traceback lines; after a preservation-first fix, the rule
scores 15.3\% savings at 99.2\% marker recall with zero critical drops.
Reducing nominal savings from 31.9\% to 15.3\% \emph{doubled} the
fidelity of the evidence the model needs; interactive task-success evaluation remained blocked by the
missing container runtime.

\paragraph{Structural retrieval (candidate quality).} On 8 repository
tasks (38 query rows) with test-grounded gold, plain
ripgrep~\citep{gallant2026ripgrep} returned a mean 310.9 matches / 2,905
tokens per query at 0.66 precision, against 170.9 / 2,090 at 0.995 for the
ast-grep hybrid; on the 14 point-definition queries the contrast is 75.6
matches / 752.8 tokens / 0.44 precision vs.\ 17.2 / 335.4 / 0.99. Under a
150-token budget, gold survival was 28.6\% (ripgrep) vs.\ 85.7\%
(structural). This demonstrates candidate-quality improvement, not
end-to-end task-success improvement: agent-level evaluation was not
performed.

\paragraph{Caveman (agent-prose compression).} On seven synthetic
fixtures, Caveman achieved 22.61\% real-token savings (721$\to$558 tokens
by provider count; 25.84\% by bytes---bytes overstate token savings) with
zero critical-marker drops. It is a constrained component study on
synthetic prose, not an end-to-end coding-agent result.

\paragraph{Embedding retrieval.} The BGE embedding leg was enabled,
health-checked, and outage-free throughout the authoritative campaign, and
its component-level union configurations improved recall-at-savings on
ContextBench slices; but no evaluated task shows a decisive
embedding-attributable effect in the gate traces. Successful activation is
not measured task-success improvement.

\paragraph{Task success at component scale.} Even where the component
harness could measure task success directly (\cref{sec:grounded}),
the evaluated compression configuration showed no positive solve-level improvement in that small single-shot study, and a
plausible-looking ratio metric (successes per million tokens) would have reversed the comparison; ratio metrics over tiny, non-overlapping success sets
are a reporting hazard, not evidence.

\begin{figure}[t]
\centering
\includegraphics[width=0.62\linewidth]{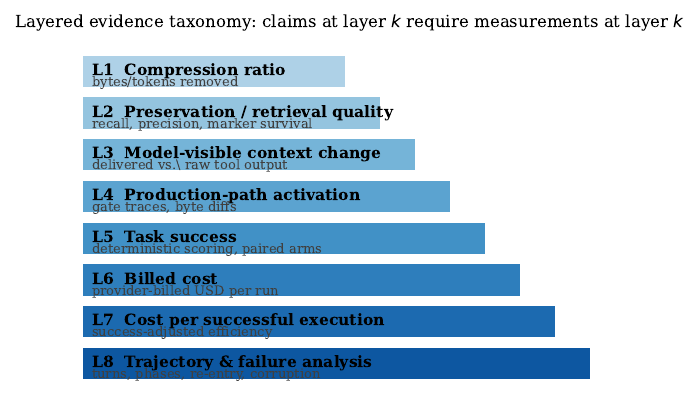}
\caption{Layered evidence taxonomy for token-efficiency claims. Every system
in \cref{tab:archmatrix} is placed at the highest layer its retained
artifacts support.}
\label{fig:layers}
\end{figure}

\paragraph{Implication.} Recall is not task success; marker
preservation is not task success; candidate precision is not task success;
synthetic critical-drop tests are not task success; bytes removed are not
billed savings; and shorter context can cause longer trajectories. Each component metric is useful for preliminary evaluation---the InterCode fix and the
stdout-consumer guard both came from component-level failures---but
claims about economic efficiency require end-to-end, success-adjusted, actually-billed evaluation. \Cref{fig:layers} summarizes the evidence taxonomy we
propose: L1 compression ratio, L2 preservation/retrieval quality, L3
model-visible context change, L4 production-path activation, L5 task
success, L6 billed cost, L7 cost per successful execution, L8
trajectory-and-failure analysis.

\section{Discussion}\label{sec:discussion}

\paragraph{Addressable cost share.} The cost composition in
\Cref{sec:composition} places an upper bound on the effect of any tool-output layer:
generated output is $\approx$11\% of the reconstructed cost
($\approx$10\% of the bill), uncached input $\approx$1\%, and the
remainder of the reconstruction ($\approx$87\%; about 80\% of the actual
bill, with the 8.7\% dollar-weighted residual unattributed) is cache
traffic priced at 10--125\% of the input rate. A layer that modifies only delivered tool text can affect only a subset of cache-read traffic, which is already discounted $10\times$. Positive end-to-end cost savings therefore require compression that does not induce offsetting trajectory changes. The largest observed cost increases (Headroom's $+48\%$ and the \ninegate{} arm's downstream cost increase) were associated with trajectory and cache effects rather than generated output alone.

\paragraph{Stage distribution versus stage cost.} In these campaigns and under
the frozen turn-level taxonomy, coarse phase-token shares were similar
across arms while phase \emph{prices} moved. This suggests evaluating interventions by phase-specific cost changes across retrieval, diagnosis, implementation, and verification, including repeated transitions into earlier phases.

\paragraph{Addressable versus framework-owned cache.} The cache-dominated
bill does not imply a large compression opportunity. In the baseline arm,
the first API call already reads 16.5k/23.2k/14.4k cached tokens
(Haiku/Sonnet/Opus) before any run content exists---a pre-existing shared
framework prefix---giving a measured lower bound of 76\%/83\%/85\% of
all cache-read volume that is framework-owned rather than user-addressable.
Delivered tool output averages only $\approx$2{,}122/417/520 estimated
tokens per run, of which $\approx$36\% (Haiku) flows through the
hook-rewritable shell route. Bounded scenarios for hook-addressable cache
dollars: conservative $\approx$4\% of cache dollars (shell-only surface),
central $\approx$12\% on Haiku ($\approx$3\% Sonnet, $\approx$5\%
Opus), upper $\approx$25--30\% (all incremental non-assistant content).
The observed results show that the \ninegate{} arm removed 38\% of
estimated shell tool-output tokens yet showed cache volumes no lower than
baseline (writes $+3\%$, reads $+4.7\%$ per run)---downstream trajectory changes offset the reduction in delivered content. The addressable share also \emph{falls}
with trajectory length (tool volume grows 1.7$\times$ from short to
7+-turn runs while cache reads grow 5.7$\times$). Exact attribution would
require per-request source labeling (system prompt, tool schemas, user
prompt, assistant history, tool output, hook-injected text, framework
metadata) for every cache write and read; that instrumentation does not
exist in the retained campaign.

\paragraph{Heterogeneity across systems.} No system dominated. RTK v0.44.1 was near cost-neutral in the aggregate campaign, with no detected success-rate
reduction at that study's precision (its byte-preserving passthrough when rules do not match also limits its maximum possible savings; and its small
pooled saving is not holdout-confirmed, \cref{tab:holdout}). The \ninegate{} stack achieved model-visible reduction but increased net cost on the aggregate workload; cost reductions occurred primarily in tasks dominated by bulk lexical retrieval and on Haiku. The tested Headroom v0.27.0 configuration
showed higher billed cost on the measured workloads without a detected
success-rate difference; the study cannot generalize to other
configurations or versions. Model and effort mattered less than task
composition; none of the arm$\times$model or arm$\times$effort
interactions survived the expansion analysis.

\paragraph{Content-type policies.} The safest observed behaviors are
content-aware: preserve-by-default for dense evidence (tracebacks, test
output, exact-byte tasks), compress redundant retrieval, never transform
streams consumed by other programs. The additional complexity of semantic rankers may be justified where lexical signals are insufficient; the structural results suggest
AST-anchored retrieval produced the strongest component-level precision result, but it
awaits end-to-end validation.

\paragraph{Association vs.\ mechanism.} Our black-box observations of
Headroom (cache-read growth concentrated in early phases, higher re-entry, the largest
billed-vs-usage residual) are associations that are consistent with several
mechanisms (payload transformation, request handling, retry behavior); we
cannot distinguish them without instrumenting the proxy. Similarly, the
\ninegate{} downstream cost increase is an observed pattern in the frozen
ledgers, not a proven causal chain.

\section{Limitations}\label{sec:limitations}

\paragraph{Authoritative campaign.} One agent harness (Claude Code) and
one provider's billing model, prices, and cache semantics at one point in
time (dollar results embed July-2026 prices; token quantities are reported
alongside). Three model families, but Sonnet 5 and Opus 4.8 were not
measured at medium/xhigh/max effort, and Opus cells are thin (8--30 paired
tasks). Baseline success rates were 96--98\%, limiting sensitivity to small differences; the study was not
designed as a formal non-inferiority trial, and no pre-specified
non-inferiority margin exists: success intervals still permit small
positive or negative differences, and small performance losses may be
undetectable. Raw (pre-layer) tool-output boundaries are unobservable for
baseline and Headroom. All within-block runs fall inside the provider's
five-minute cache TTL, so cross-arm cache carryover cannot be excluded by
temporal separation (\cref{sec:carryover}); randomization balance and null
order-interactions are the available evidence. The cost reconstruction
leaves an 8.7\% dollar-weighted aggregate residual, with per-model usage
for possible secondary calls not persisted. Judges are deterministic
scripts; graded quality is not measured.

\paragraph{Thinking tokens.} Thinking tokens were not exposed separately
by the API and telemetry used in these campaigns and therefore cannot be
reconstructed from the retained artifacts; reasoning-token usage was not
separately observable; generated output and observable diagnosis-phase
behavior provide only partial behavioral indicators and are not direct
measurements of latent reasoning. Post-hoc evidence
(\cref{sec:calibration}) indicates the exposure differs by model:
Sonnet~5's aggregate usage reconciles with its bill ($+0.2\%$), while
Haiku~4.5 carries an effort-scaling, thinking-consistent charge inside the
disclosed residual; cross-model decompositions should not assume uniform
usage-field semantics.

\paragraph{Headroom.} One tested version (0.27.0) and one configuration
(token mode, caching enabled, code-aware disabled); black-box internals;
raw-vs-delivered output unobservable; the largest unexplained billing
residual ($+14.5\%$). Results cannot be generalized to other
configurations or versions.

\paragraph{Vendor CLI compressor and \texttt{ml\_lexical}.} The vendor
CLI compressor has no same-harness end-to-end evidence (vendor-claimed
figures only). The \texttt{ml\_lexical} engine has component-level
retrieval evidence but no billed task-success study.

\paragraph{Caveman.} Seven synthetic fixtures; no production wiring; no
real-traffic evaluation.

\paragraph{Grounded single-shot study.} One model (Sonnet 4.6); small $n$
(solve counts of five and two; exact paired $p=0.25$); no recovery loop;
SEARCH/REPLACE-specific edit application; SWE-bench-derived Go tasks; one
compression configuration; per-row costs not retained (no uncertainty for the per-resolved-row
cost); the IC-SWE pilot's cost is trace-derived.

\paragraph{Phase analysis.} Deterministic rules validated by an
implementation audit, not by blind semantic annotation; mutually exclusive
labels force mixed turns into single categories; planning is structurally
undercounted; base-campaign splits only.

\paragraph{Campaign composition.} The long-session campaign used a
different experimental configuration and is reported separately throughout. The
task mix is weighted toward retrieval, navigation, and debugging;
generalization beyond Claude Code and the tested workloads is not
established. All RTK experiments reported in this paper use v0.44.1. A post-hoc paired
replication on additional Opus medium-effort cells reproduced the aggregate
near-null pattern (\cref{app:upstream}), though on a half-size task sample
and a newer harness version (Claude Code 2.1.220).

\paragraph{Interactive sessions (unmeasured extrapolation).} All
end-to-end dollar results come from single-prompt benchmark runs, whose
context is dominated by the framework prefix: the user-accessible surface
(tool outputs, tool-call arguments, retrieved files, injected context,
conversation history) is $\approx$6\% of cost, which places the estimated upper bound for a user-side compression layer at $\approx$5\% of cost under the measured decomposition---and the share of that surface
the evaluated layers actually captured makes the estimated realized effect smaller ($\approx$30\% capture on the modified surface). A companion
composition analysis of 41 interactive developer sessions (3{,}991 API
requests, reconstructed from retained Claude Code telemetry under the
same tokenizer calibration) finds a materially different mix: the
framework base falls to $\approx$8\% of input tokens and the
user-accessible surface grows to roughly 30\% of modeled cost, dominated by
tool-call arguments ($\approx$26\% of input), injected
\texttt{CLAUDE.md}/reminder text ($\approx$10\%), and bulkier tool
outputs and file reads. Holding the benchmark capture rate constant, the
same arithmetic scales the realized-savings bound from $\approx$1.5\% to
$\approx$9\% of modeled cost in interactive sessions. We report this as an
upper bound on an \emph{unmeasured} setting, not a forecast: the
countervailing effects measured in this paper (added thinking, added
turns) can grow with the same surface, no per-session ROI instrumentation
exists for interactive conversations, and the capture-rate transfer is an
assumption. Measuring this setting directly---with per-session ROI instrumentation and an empirical interactive capture rate---is a high-priority follow-up, alongside isolated measurement of tool-call arguments,
end-to-end evaluation of retrieval and conversation-history compression,
and the Codex distributional analysis noted in \cref{sec:codex}.

\section{Recommendations for Benchmark Design}\label{sec:recommendations}

A trustworthy token-efficiency benchmark for coding agents should report,
per paired task: task success; actual billed cost; cache-creation,
cache-read, uncached-input, and generated tokens; turns and tool calls;
raw and delivered tool volume where observable; retries and failures with
exclusion policy; production-path activation proof; model and effort; task
family; cost per successful execution; and confidence intervals with the
clustering unit stated. Aggregates that omit any of these can lead to
failure modes documented above (early-failure ``savings'', inactive-path
comparisons, trajectory-blind ratios).

Because content types fail differently, we recommend separate tracks
rather than one blended score: (i) redundant bulk retrieval; (ii)
long-history memory; (iii) dense diagnostic evidence (tests, tracebacks,
logs) with preservation gates; (iv) structural code search; and (v) full
end-to-end software tasks with success-adjusted billing as the terminal
metric. Component tracks (i)--(iv) remain essential as screens and for
failure analysis---they found both safety bugs reported here---but claims
about \emph{efficiency} should be licensed only by track (v).

\section{Related Work}\label{sec:related}

\paragraph{Coding-agent benchmarks.} SWE-bench and its verified subset
\citep{jimenez2024swebench,openai2024swebenchverified} established
repository-level issue resolution as the standard end-to-end coding task;
InterCode \citep{yang2023intercode} standardized interactive coding with
execution feedback; HumanEval and MBPP
\citep{chen2021codex,austin2021mbpp} anchor function-level synthesis.
Agent harnesses around these benchmarks
\citep{yang2024sweagent,wang2024openhands,wang2024codeact,yao2023react}
report success and sometimes token counts, but rarely billed cost with
cache decomposition; $\tau$-bench \citep{yao2024taubench} adds
user-interaction realism. Our contribution is complementary: a cost-composition
and evidence-standard study over paired arms of the same harness. Many prior systems are primarily evaluated with component-level metrics,
while fewer report paired task success, provider-returned usage,
trajectory length, and success-adjusted dollar cost together; \cref{tab:relwork} maps representative prior
work onto the evidence layers of \cref{sec:componentgap}.

\begin{table}[t]
\centering
\caption{Evidence dimensions reported by representative prior work vs.\
this study. Many prior systems are primarily evaluated with component-level
metrics; entries reflect the cited papers' primary evaluations, not all
follow-up work. TS = task success.}
\label{tab:relwork}
\small
\setlength{\tabcolsep}{3pt}
\resizebox{\textwidth}{!}{%
\begin{tabular}{lcccccc}
\toprule
Work & Compression/recall & Model-visible ctx & TS & Cost evidence & Trajectory & Success-adj.\ cost \\
\midrule
LLMLingua / LongLLMLingua~\citep{jiang2023llmlingua,jiang2024longllmlingua} & \checkmark & \checkmark & QA acc. & --- & --- & --- \\
Selective Context~\citep{li2023selective} & \checkmark & \checkmark & QA acc. & --- & --- & --- \\
Gist tokens~\citep{mu2023gist} & \checkmark & \checkmark & --- & --- & --- & --- \\
RepoCoder~\citep{zhang2023repocoder} & recall & \checkmark & completion & --- & --- & --- \\
SWE-bench / SWE-agent~\citep{jimenez2024swebench,yang2024sweagent} & --- & --- & \checkmark & --- & partial & --- \\
InterCode~\citep{yang2023intercode} & --- & --- & \checkmark & --- & turns & --- \\
$\tau$-bench~\citep{yao2024taubench} & --- & --- & \checkmark & --- & \checkmark & --- \\
FrugalGPT~\citep{chen2023frugalgpt} & --- & --- & \checkmark & API prices & --- & partial \\
Prompt Cache / vLLM~\citep{gim2024promptcache,kwon2023vllm} & --- & --- & --- & latency only & --- & --- \\
\textbf{This study} & \checkmark & \checkmark & \checkmark & billed+usage & \checkmark & \checkmark \\
\bottomrule
\end{tabular}
}
\end{table}

\paragraph{Long context and retrieval.} Long-context evaluation shows that
models use long inputs unevenly \citep{liu2024lost,bai2024longbench}, and
that retrieval remains competitive with long-context reading
\citep{xu2024retrievallong}; retrieval-augmented generation
\citep{lewis2020rag} and repository-level retrieval for code
\citep{zhang2023repocoder,husain2019codesearchnet,feng2020codebert}
motivate the ranking legs studied here, with BGE embeddings
\citep{xiao2023bge} as the semantic backbone and ast-grep
\citep{astgrep2026} for structural search.

\paragraph{Context compression.} Prompt- and context-compression methods
\citep{jiang2023llmlingua,jiang2024longllmlingua,li2023selective,mu2023gist}
optimize token counts at fixed quality, typically measured by downstream QA
accuracy rather than billed agent cost. Our results argue that for agents,
such component gains must additionally clear cache-pricing and trajectory
effects before they become savings.

\paragraph{Serving economics and cost-aware inference.} Prompt caching and
attention reuse \citep{gim2024promptcache,kwon2023vllm} changed the price
structure this paper measures; provider documentation specifies the billed
multipliers \citep{anthropic2026caching,anthropic2026pricing}. Cost-aware
cascades and routing \citep{chen2023frugalgpt} and the efficiency survey of
\citet{wan2024efficiency} study model choice as an optimization variable; we study context policy as an optimization variable and show that its billed effect is dominated by cache and trajectory terms. Tool-use foundations \citep{schick2023toolformer}
underlie the agent loop whose economics we measure.

\section{Conclusion}\label{sec:conclusion}

Token compression and cost reduction are not interchangeable quantities.
In 2,848 paired, provider-billed coding-agent runs, cache creation and
cache reads dominated the reconstructed cost composition ($\approx$87\% of
the four-component reconstruction; $\approx$80\% of the actual bill, with
an explicitly quantified 8.7\% unattributed residual), generated output
was a minority share, and removing an estimated 38\% of raw tool output
coexisted with
a $+6.8\%$ paired cost increase as trajectory changes---extra diagnosis, testing, and re-retrieval turns---offset the retrieval-stage savings. Component benchmarks were indispensable for finding safety
failures (traceback truncation, shell-pipeline incompatibility) but did not
predict end-to-end outcomes; per-task tool-output reduction was a weak, unstable predictor of
billed-cost change. Effective context policy in these
workloads is task-, content-, model-, and effort-aware: preserve dense
evidence, compress redundant retrieval, never transform program-consumed
streams, and evaluate every claim at the level that matters---end-to-end,
success-adjusted, actually-billed cost. A cross-agent replication provides additional evidence: the same Headroom family that
showed the largest cost increase on Claude Code (CPS ratio 1.464) was
associated with a 12--14\% reconstructed-cost \emph{saving} at equal
success on Codex, with zero compressor activity in both settings and a
tripled wall time (\cref{sec:codex}). Efficiency-layer effects are
properties of the layer--harness--model--workload combination.
No system we measured is universally best, and we make no such claim. The principal methodological contribution is the proposed measurement standard.

\section*{Disclosure}
The authors are affiliated with PointFive. The authors are not the
developers of Headroom. All arms were built and configured by the authors:
the evaluated hook-based layers were compiled from source by the authors
(Appendix~\ref{app:provenance} records the commits and binary hashes), and
Headroom ran its open-source distribution with default proxy settings
(token mode, caching enabled, code-aware mode disabled) with no vendor
guidance; no external vendor was given an opportunity to verify any
configuration. To reduce evaluation bias, the study uses frozen task
manifests, symmetric deterministic judges, randomized arm order, retained
transcripts, append-only ledgers, and released analysis artifacts; these
controls reduce but do not eliminate the risks inherent in
author-configured evaluation.

\section*{Artifact Availability}
This paper is accompanied by a benchmark-and-data release (a partial
artifact release): the benchmark harness and deterministic judges, frozen
task manifests and pre-specification documents, arm and run provenance
(including pinned binary hashes), fixture builders with pinned repository
SHAs, the campaign analysis scripts, and the derived run-level datasets
(per-run metadata with provider usage and billed cost, the append-only
cost ledger, per-turn and per-tool-call ledgers, phase assignments, and
machine-readable statistical outputs). Raw transcripts are withheld
because they embed third-party repository content and proxied system
payloads; derived per-turn ledgers are released in their place, so the
statistical results are computationally reproducible from the release
while raw-trace re-audits are not. The release contains no paper text or
result descriptions; the paper's \LaTeX{} sources and its figure/table
generation scripts accompany the arXiv submission. The release is
available at
\url{https://github.com/PointFiveLabs/ai-efficiency-benchmark}
(immutable release tag \texttt{v1.0.0}), licensed under CC~BY~4.0. The Codex replication artifacts (\cref{sec:codex}) are retained in their
own campaign tree and are not part of this release. We do not claim full
reproducibility of the campaigns themselves, which would require
re-running paid API workloads.

\bibliographystyle{plainnat}
\bibliography{references}

\clearpage
\appendix

\section{Reproducibility Appendix}\label{app:repro}

\subsection{Artifact map and campaign hierarchy}\label{app:artifacts}

The study is based on retained transcripts, ledgers, manifests, gate
traces, source code, and generated analysis files; no paid runs were
executed for the paper-writing audit. The evidence index is the audit
directory \path{research/token_efficiency_paper_gap_audit/} (fifteen
deliverables plus \texttt{gap\_evidence.json},
\texttt{composition\_data.json}, \texttt{family\_matrix.json},
\texttt{phase\_shares.json}, and four evidence-miner reports). The
authoritative campaign lives in
\path{research/benchmark_validation_campaign_20260707/} with
\path{deliverables/EXPANSION_FINAL_REPORT.md} as its final report;
its raw ledgers are \texttt{results.jsonl} (2,977 rows),
\texttt{ledger\_turns.jsonl} (13,620), \path{ledger_tool_calls.jsonl}
(7,097), \path{phase_assignments.jsonl} (13,620),
\texttt{COST\_LEDGER.csv}, and the frozen manifests. Earlier campaigns:
\path{research/full_benchmark_campaign_20260705_150132_full/}
(Stage~0), \path{research/full_benchmark_campaign_20260701_133214/}
(Stages~1--2), and the early-generation campaigns and long-session
campaign under the companion research tree's
\texttt{benchmark\_comparison/} directory. Component benchmarks and the single-shot grounding program live in the
benchmark worktree (\texttt{cli/} harnesses; \path{outputs/agent_pilot/},
\path{outputs/ast_retrieval/}, \path{outputs/intercode_proxy/}) and the
companion outputs tree (\path{outputs/contextbench/} sweeps,
\path{outputs/locoeval/}, \path{outputs/caveman/}, the
\path{cb_e2e_*} phase reports, \path{benchmark_sweep_2026-06-07.md},
\path{TEST_RESULTS_SUMMARY.md}, and
\path{contextbench_agent_smoke.md}); their extraction into this paper is
\path{scripts/extract_bench_program.py} $\to$
\path{data/bench_program.json}.

\subsection{Campaign reconciliation}\label{app:reconciliation}

\begin{table}[h]
\centering
\caption{Exact construction of the global execution and spend totals, by
evidence class.}
\small
\resizebox{\textwidth}{!}{%
\begin{tabular}{lcp{7.2cm}}
\toprule
Evidence class & Executions & Basis \\
\midrule
Runtime cost ledgers & 5,123 & append-only per-run ledgers (validation, Stages 1--2 incl.\ rerun, pilot, comprehensive, three long-session campaigns, expanded API) \\
Suite-results derived & 8 & costsmoke: 4 tools $\times$ 2 cases from the suite results file (no per-run ledger) \\
Report/trace derived & 362 & single-shot grounding program (frozen phase reports; IC-SWE pilot cost derived from trace token fields) \\
\midrule
\textbf{Billed total} & \textbf{5,493} & \\
Component evaluations & 8,263 & token-counting / estimator basis, no billed inference \\
Deterministic checks & 4 & Stage-0 gate byte-diff proof \\
\bottomrule
\end{tabular}
}
\end{table}

The costsmoke comparison is an independent early campaign (not a subset or
duplicate of any other row); its spend is recorded in its suite
results file, and its execution count (4 tools $\times$ 2 cases) is derived
from that file rather than from a runtime ledger. The IC-SWE pilot's
cost is derived from per-turn token fields in its trajectory
traces at Sonnet 4.6 prices. All other totals sum append-only per-run
ledger rows.

\subsection{Normalized four-component shares}\label{app:normshares}

The table below restates the cost composition over the four-component
reconstruction denominator (the one used by earlier drafts), excluding the
billed residual.

\begin{table}[H]
\centering
\caption{Normalized four-component shares (\% of the reconstructed
$\widehat{C}$, excluding the billed residual), by slice --- the
denominator used by earlier drafts and by finer slices (models, efforts,
sessions, outcomes, families). Billed-basis shares including the residual
are in \cref{tab:coststack}.}
\small
\resizebox{\textwidth}{!}{%
\begin{tabular}{lcccc}
\toprule
Slice & {Uncached input} & {Cache write} & {Cache read} & {Generated output} \\
\midrule
All runs & 1.4 [1.2, 1.6] & 48.5 [47.6, 49.3] & 38.7 [38.0, 39.5] & 11.4 [11.0, 11.9] \\
Baseline & 1.5 [1.1, 2.0] & 48.1 [46.6, 49.6] & 37.8 [36.6, 39.0] & 12.6 [11.6, 13.6] \\
RTK & 1.6 [1.2, 2.1] & 49.1 [47.7, 50.4] & 37.0 [36.0, 38.1] & 12.2 [11.4, 13.1] \\
RTK-ML & 1.5 [1.1, 1.9] & 46.3 [44.5, 48.1] & 39.0 [37.4, 40.5] & 13.2 [12.3, 14.2] \\
Headroom & 1.0 [0.7, 1.3] & 50.0 [48.1, 52.1] & 40.4 [38.8, 42.1] & 8.5 [7.9, 9.2] \\
Successful runs & 1.4 [1.2, 1.6] & 48.7 [47.8, 49.6] & 38.6 [37.9, 39.3] & 11.3 [10.9, 11.8] \\
Failed runs & 0.7 [0.0, 1.6] & 41.7 [36.2, 48.3] & 43.5 [38.7, 47.7] & 14.1 [12.0, 15.8] \\
Short sessions & 1.6 [1.3, 1.9] & 56.8 [55.8, 57.8] & 34.1 [33.4, 34.9] & 7.4 [7.1, 7.8] \\
Long sessions & 1.6 [1.1, 2.2] & 38.8 [36.7, 40.8] & 41.1 [38.9, 43.5] & 18.4 [17.0, 19.7] \\
\bottomrule
\end{tabular}
}
\end{table}

\subsection{One- and two-task families (exploratory)}\label{app:familyfull}

The table below lists the exploratory family cells excluded from the
main-text family comparison.

\begin{table}[H]
\centering
\caption{Task families with fewer than three analyzed tasks (descriptive
only; no CIs; do not compare against the well-powered families of
\cref{tab:family}).}
\small
\begin{tabular}{lrccc}
\toprule
Task family & $n_T$ & RTK $\Delta$\% & RTK-ML $\Delta$\% & Headroom $\Delta$\% \\
\midrule
doc\_grounded\_impl & 2 & +4.2\textsuperscript{s} & +13.0\textsuperscript{s} & +40.0\textsuperscript{s} \\
exact\_preservation & 2 & -2.0\textsuperscript{s} & -3.6\textsuperscript{s} & +61.6\textsuperscript{s} \\
regression\_investigation & 2 & +4.2\textsuperscript{s} & +5.0\textsuperscript{s} & +36.3\textsuperscript{s} \\
code\_review & 1 & -1.8\textsuperscript{s} & -7.2\textsuperscript{s} & +14.7\textsuperscript{s} \\
config\_dependency & 1 & -8.5\textsuperscript{s} & +50.8\textsuperscript{s} & +30.4\textsuperscript{s} \\
localized\_debug & 1 & -3.2\textsuperscript{s} & +28.9\textsuperscript{s} & +30.7\textsuperscript{s} \\
migration & 1 & -17.2\textsuperscript{s} & +12.1\textsuperscript{s} & +16.4\textsuperscript{s} \\
performance & 1 & +3.6\textsuperscript{s} & +32.3\textsuperscript{s} & +82.4\textsuperscript{s} \\
security\_analysis & 1 & +9.7\textsuperscript{s} & -4.7\textsuperscript{s} & +56.5\textsuperscript{s} \\
shell\_workflow & 1 & -1.6\textsuperscript{s} & -0.9\textsuperscript{s} & +34.1\textsuperscript{s} \\
test\_repair & 1 & -2.2\textsuperscript{s} & +22.7\textsuperscript{s} & +30.7\textsuperscript{s} \\
\bottomrule
\end{tabular}
\end{table}

\subsection{Result-to-claim traceability}\label{app:traceability}

The table below maps each primary claim to its artifact, script,
population, and provenance class.

\begin{table}[H]
\centering
\caption{Primary claims mapped to artifacts, scripts, populations, and
provenance class (PS = pre-specified analysis on the frozen campaign; HC =
holdout-confirmed; PL = pooled; EX = exploratory; DS = descriptive; RP =
transcribed from a frozen external replication report).}
\small
\setlength{\tabcolsep}{3pt}
\resizebox{\textwidth}{!}{%
\begin{tabular}{lllll}
\toprule
Claim & Data / script & Population & CI method & Class \\
\midrule
Billed-basis cost shares + residual & revision\_analyses r1 / extract\_revision\_analyses.py & 2,848 runs & run bootstrap (2k, seed 7) & PS/DS \\
8.7\% aggregate residual & revision\_analyses r1 & 2,848 runs & run bootstrap & DS \\
$+6.8\%$ RTK-ML cost & paper\_data paired\_overall / extract\_paper\_data.py & 103 tasks & task bootstrap (10k, seed 7) & PL (HC: $+9.1\%$) \\
$-2.7\%$ RTK cost & paper\_data paired\_overall & 103 tasks & task bootstrap & PL only (holdout CI crosses 0) \\
$+48.4\%$ Headroom cost & paper\_data paired\_overall & 103 tasks & task bootstrap & PL (HC: $+47.3\%$) \\
Success differences (CIs cross 0) & paper\_data paired\_overall & 103 tasks & task bootstrap & PS \\
38.4\% estimated raw tool-output reduction (local BPE tokenizer) & paper\_data aggregate\_by\_arm & RTK-arm ledger & --- & DS \\
$r=0.154$ reduction--cost & revision\_analyses r4 & 100 Haiku tasks & task bootstrap & EX/DS \\
Applies 27/40 vs 15/40 & bench\_program (Phase E report) & 40 rows $\times$ 2 & exact counts & DS \\
Grounded 5/29 vs 2/29 & revision\_analyses r6 & 29 rows & exact paired $p=0.25$ & DS \\
Phase shares & paper\_data phase\_shares & 10,376 turns & --- & DS \\
Re-entry / first-edit deltas & phase ledger reports & base splits & --- & DS \\
Pipeline-incompatibility failures (4/4) & stage1 failure report & stage-1 & exact counts & DS \\
Cost-per-success ratios & revision\_analyses r2 & 103 tasks & task bootstrap & PS \\
Order-sensitivity nulls & revision\_analyses r3 & 2,848 runs & task bootstrap & PS (sensitivity) \\
Haiku effort-scaling residual (thinking-consistent) & thinking\_addressability / extract\_thinking\_addressability.py & non-Headroom runs by effort & --- & DS/EX \\
$\geq$76--85\% framework-prefix share of cache reads & thinking\_addressability & baseline runs + turn ledger & --- & DS (lower bound) \\
$-12.49\%$ Codex Headroom cost, 39/40 both arms & Codex replication report (external tree) & 40 disjoint task pairs & --- & RP \\
$-14.35\%$ combined 52-pair Codex difference & Codex replication report (external tree) & 52 task pairs & --- & RP (descriptive pooling) \\
\bottomrule
\end{tabular}
}
\end{table}

\subsection{Coverage, formulas, and unavailable fields}\label{app:coverage}

Model$\times$effort coverage of the main campaign is exactly
nine cells of \Cref{tab:modeleffort}. The post-hoc replication campaign (\cref{app:upstream}) adds two
medium-effort Opus cells for RTK v0.44.1; those same cells are also reported
in \cref{tab:cps,tab:upstream}. Sonnet~5 at medium/xhigh/max and
Opus~4.8 at xhigh/max were never
measured. Billed cost is provider \texttt{total\_cost\_usd};
reconstruction follows \cref{eq:decomp} with the published per-model
list prices, $\mu_{\mathrm{r}}{=}0.1$,
$\mu_{\mathrm{w}}{=}1.25$; the calibration table (standard vs.\
introductory prices, 5-minute vs.\ 1-hour multipliers) is in the audit's
\path{COST_FIELDS_AND_FORMULAS.md}. Fields unavailable in the
retained artifacts: separately-exposed thinking tokens; per-tool-call
dollars; per-model usage splits within a run; raw tool-output boundaries
for baseline/Headroom; time-to-first-correct-file; Stage-1 transcripts
(deleted post-harvest); expansion-split phase ledgers (regenerable from
retained gzipped transcripts).

\subsection{Exclusion policy}\label{app:exclusions}

Billing totals (\cref{tab:campaigns}) count every ledger row (each row is
one billed execution); analysis populations apply, separately: keep-last
dedupe per \texttt{run\_id}; require
\texttt{status=completed} and no infrastructure failure; drop the
calibration split; drop the one manifest-flagged defective task
(a holdout refactor whose file constraints contradicted its edit
requirements); judge-defect fixes applied symmetrically by rescoring
stored outputs, with append-only history. Result: 2,848 analyzed of 2,908
executed runs.

\subsection{System provenance}\label{app:provenance}

\begin{table}[h]
\centering
\caption{System versions and provenance (authoritative campaign).}
\small
\begin{tabular}{llp{6.2cm}}
\toprule
System & Version & Provenance \\
\midrule
Claude Code & 2.1.201 & official extension native binary \\
RTK & v0.44.1 & open-source RTK distribution (\url{https://github.com/rtk-ai/rtk}); same version used throughout the reported RTK experiments \\
\Ninegate & guarded release build & authors' branch build; all gate flags on; safety router enforce \\
Headroom & 0.27.0 & pip distribution; \texttt{ANTHROPIC\_BASE\_URL} proxy \\
Embedding sidecar & bge-small-en-v1.5 & offline HF cache (\texttt{HF\_HUB\_OFFLINE=1}); loopback only \\
\bottomrule
\end{tabular}
\end{table}

Gate activation: routes/query-propagations/skips of 521/916/339 (Haiku),
208/395/150 (Sonnet), 26/57/28 (Opus) on the \ninegate{} arm and exactly
zero on all comparator arms; free-fixture byte proof 11,050\,B
$\to$2,570\,B with gates on, byte-identical with gates off, and
byte-identical with all gate variables set on RTK (gate-env inert).

\subsection{Dataset snapshots}\label{app:snapshots}

Frozen: task manifests (45 base + expansion; 103 analyzed), six repository
pin files (click \texttt{16fc00e2\ldots}, cobra \texttt{ad460ea8\ldots},
express \texttt{ba006766\ldots}, flask \texttt{36e4a824\ldots}, gin
\texttt{34dac209\ldots}, requests \texttt{4c800e9a\ldots}), seeded
synthetic monorepo builders, gzipped transcripts for every authoritative
run. Weaker: the ContextBench Hugging Face revision hash was not recorded
(n=10 row identifiers retained: 445, 4, 138, 210, 135, 324, 307, 492,
454, 198); LoCoEval hop files were fetched on demand; the 50-row preservation results
are retained in the companion outputs tree while some intermediate
artifacts resided in temporary directories; several component
reports cross-reference a sibling research tree by absolute path.

\subsection{Statistical procedure and regeneration}\label{app:regen}

Statistics follow \cref{sec:stats} (task-clustered pairing; 10,000
bootstrap resamples, seed 7; repository-clustered and
leave-one-repository-out variants; exploratory family screening by
CI indicators (no $p$-values, no FDR control claimed); ICC/Kish
effective sample sizes; $n<6$ cells flagged). All tables and figures in
this paper regenerate offline with:

\begin{quote}\ttfamily
python3 scripts/extract\_paper\_data.py\\
python3 scripts/extract\_bench\_program.py\\
python3 scripts/extract\_thinking\_addressability.py\\
python3 scripts/make\_tables.py\\
python3 scripts/make\_figures.py\\
make paper
\end{quote}

\subsection{Integrity statement}\label{app:integrity}

No tracked production file was modified during the audit. The generated
research outputs were written under the untracked research directory.
Pre-existing untracked directories require separate session-start evidence
and are not themselves proof of audit modifications. No paid API
experiments were run for the audit or for this paper; every quantitative
statement derives from retained artifacts of the completed campaigns.

\subsection{Created benchmark corpora and synthetic fixtures}\label{app:corpora}

\begin{table}[h]
\centering
\caption{Benchmark suites run and synthetic corpora created by the
measurement program (companion research trees; counts from the retained
artifacts).}
\small
\setlength{\tabcolsep}{4pt}
\resizebox{\textwidth}{!}{%
\begin{tabular}{llp{6.8cm}}
\toprule
Corpus / suite & Size & Role \\
\midrule
Pilot task fixtures (\path{benchmark_comparison/e2e/fixtures}) & 7 tasks, 17 files & scripted bugfix, navigation, log, feature, refactor, regression, and retrieval tasks \\
Comprehensive fixtures & 14 suites, 343 files & pass/fail test suites (40--400 tests), bulk-output and grep-tree generators \\
Long-session repositories & 3 repos, 911 files & synthetic large repos (small/std/large) for marathon sessions \\
Stage 1--2 seeded fixtures & in-code + monorepo & deterministic builders, committed ground truth \\
Authoritative synthetic monorepo & seeded, 2 builders & \path{build_synth.py} + extension; ground-truth JSON \\
cmdcompress corpus & 159 cases (11 families) & command-output compression at scale; 34 live-API cases \\
Output-compression cases & 113 files & observer vs.\ RTK behavioral cases + runtime proof \\
privacybench / observer\_intent / retrieval-modes / go-retrieval & suite-level & redaction, intent top-1, retrieval-mode exercisers \\
InterCode captured corpus & 24 observations (240 rows) & real pytest tracebacks, 12 MBPP tasks $\times$ pass/bug \\
Caveman fixtures & 7 synthetic + 12-row control & agent-prose compression skill measurement \\
ContextBench sweep grid & 36 runs, 7{,}510 rows & top-$k$/top-$p$/adaptive/matrix policy sweeps \\
Benchmark manifests & 8 + 21 + 3 + 103 tasks & frozen task definitions across campaign generations \\
\bottomrule
\end{tabular}
}
\end{table}

\section{Reconciliation of the Headroom 18/21 Record}\label{sec:headroom1821}

An early 21-task pilot produced an apparent contradiction between
``18/21'' and ``21/21'' for Headroom. The two values refer to different outcomes:
18/21 is Headroom's \emph{task-success} count in that pilot; 21/21 is its
\emph{runs-completed} count in the same campaign---the same denominator
with two different numerators. No retained source claims 21/21 task
success. In that pilot (single model, one repetition), the Headroom arm
completed 21/21 runs and passed 18/21 tasks against the baseline's 10/21;
the source report itself flags this as a candidate proxy confound
requiring replication, since an API-boundary proxy can also alter request
handling. The subsequent authoritative campaign did not reproduce a
success advantage (success deltas' CIs cross zero;
\Cref{tab:aggregate}) and established a per-task cost increase positive in
all 23 analyzed task families ($+14.7\%$ to $+82.4\%$;
\Cref{tab:family}) and in every observed model--effort cell ($+9.0\%$ to
$+110.2\%$; \cref{tab:modeleffort}).

\section{Post-Hoc Opus-Medium Replication: RTK v0.44.1}\label{app:upstream}

After the main campaign was frozen, we ran an additional paired replication
using the same RTK version reported throughout this paper, v0.44.1, to cover
two model--effort cells absent from the main design: Claude Opus 4.8 and
Claude Opus 5 at medium effort. The purpose of this replication is therefore
coverage, not a comparison between RTK versions.

\paragraph{Design.} Baseline and RTK v0.44.1 were compared on 52 of the 103
analyzed tasks, drawn by family-stratified round-robin over all 23 analyzed
families (seed 20260730), with unchanged frozen manifests and deterministic
judges. Both models were run at \emph{medium} effort. The comparison comprises
104 paired task--model blocks (52 per model; 208 baseline/RTK runs used in the
reported pairwise analysis), all completed with zero infrastructure failures.
The harness was Claude Code 2.1.220 rather than 2.1.201, constant across the
reported pair, and the same per-run isolation protocol was used: fresh working
copies, per-run fake \texttt{\$HOME}, scrubbed environment, project-only
setting sources, and per-run budget caps. Hook activation is directly
observable in the retained transcripts: RTK rewrote commands in 68 of its 104
runs; the remainder issued no rewrite-eligible shell commands.

\begin{table}[H]
\centering
\caption{Post-hoc RTK v0.44.1 Opus-medium replication (104 paired blocks,
52 tasks; 208 baseline/RTK runs in the reported pairwise analysis). Billed
cost; paired per-task change vs.\ baseline as \% of the baseline mean,
task-clustered bootstrap 95\% CIs, 10{,}000 resamples.}
\label{tab:upstream}
\small
\begin{tabular}{llcccc}
\toprule
Arm & Model & $n$ pairs & $\Delta$cost vs.\ baseline & Success (arm/base) & CPS ratio \\
\midrule
RTK v0.44.1 & Opus 4.8 & 52 & $-2.9\%$ \ci{-18.8}{+9.3} & 98.1\% / 94.2\% & 0.933 \ci{0.795}{1.073} \\
RTK v0.44.1 & Opus 5   & 52 & $+2.5\%$ \ci{-6.1}{+12.0} & 98.1\% / 98.1\% & 1.025 \ci{0.941}{1.123} \\
RTK v0.44.1 & pooled   & 104 & $-0.1\%$ \ci{-9.3}{+7.8} & 98.1\% / 96.2\% & 0.979 \ci{0.896}{1.066} \\
\bottomrule
\end{tabular}
\end{table}

\paragraph{Result.} The two models have opposite point estimates: RTK v0.44.1
is $-2.9\%$ \ci{-18.8}{+9.3} on Opus~4.8 and $+2.5\%$
\ci{-6.1}{+12.0} on Opus~5. Pooled across the two medium-effort cells, the
paired billed-cost change is $-0.1\%$ (CI \ci{-9.3}{+7.8}) with CPS ratio
0.979 \ci{0.896}{1.066}. All intervals cross the null. These results extend
the main RTK v0.44.1 evidence to previously unmeasured Opus medium-effort
cells; they do not support a model-independent cost saving. Replication
artifacts (runner, per-run results, cost ledger, and machine-readable
statistics) are retained in the replication campaign tree.

\section{Evidence Boundaries by System}\label{app:boundaries}

\begin{table}[h]
\centering
\caption{Publishable vs.\ unsupported claims by system (condensed from the
audit's evidence-boundary and unpublishable-claims deliverables).}
\small
\setlength{\tabcolsep}{4pt}
\resizebox{\textwidth}{!}{%
\begin{tabular}{lp{5.6cm}p{5.6cm}}
\toprule
System & Supported by retained evidence & Not supported \\
\midrule
Baseline / RTK / \ninegate{} / Headroom & paired billed cost, success, trajectories, activation & causal internals for Headroom; universal rankings \\
Vendor CLI compressor & architectural description; vendor-claimed category savings, labeled as claims & any measured E2E, cost, or success figure \\
\texttt{ml\_lexical} engine & implemented architecture; component retrieval quality with real token counts & E2E benefit, task success, production traffic \\
Caveman & 22.6\% real-token savings, 0 critical drops, $n{=}7$ synthetic fixtures & production impact, real traffic, task success \\
Embedding leg & activation, health, component recall configurations & measured task-success improvement \\
Structural retrieval & candidate precision/token/gold-survival gains & end-to-end task-success improvement \\
ContextBench/LoCoEval & preservation-proxy results with stated caveats & prediction of agent behavior or billed savings \\
\bottomrule
\end{tabular}
}
\end{table}

\begin{table}[h]
\centering
\caption{Reproducibility matrix (condensed).}
\small
\begin{tabular}{lp{4.5cm}l}
\toprule
Control / gap & Detail & Status \\
\midrule
Env scrubbing, fake \texttt{\$HOME}, project-only settings & per-run isolation in all paid campaigns & in place \\
Binary sha256 pins + frozen manifests & arms and tasks & in place \\
Append-only rescoring + dedupe loaders & audit trail retained & in place \\
Offline regeneration of all statistics & frozen analysis scripts over shipped ledgers & in place \\
Provider nondeterminism & mitigated by repetitions, ICC/ESS reported & inherent \\
Stage-1 transcripts & deleted post-harvest & permanent gap \\
Pricing drift & dollars embed July-2026 prices; tokens reported alongside & inherent \\
Expansion-split phase ledgers & regenerable from retained transcripts & pending \\
Cross-tree absolute paths in component reports & sibling research tree & documented \\
\bottomrule
\end{tabular}
\end{table}

\end{document}